\pdfoutput=1 
\documentclass{article}

\usepackage{microtype}
\usepackage{graphicx}
\usepackage{subfigure}
\usepackage{booktabs} 
\usepackage{grffile} 

\usepackage{hyperref}

\usepackage[accepted]{icml2024}

\usepackage{amsmath}
\usepackage{amssymb}
\usepackage{mathtools}
\usepackage{amsthm}
\usepackage{enumitem}
\usepackage{scrextend}
\usepackage[most]{tcolorbox}
\usepackage{caption}
\usepackage[normalem]{ulem}
\useunder{\uline}{\ul}{}
\usepackage{multirow}
\usepackage{longtable}

\usepackage[capitalize,noabbrev]{cleveref}

\theoremstyle{plain}

\theoremstyle{definition}

\theoremstyle{remark}

\usepackage[textsize=tiny]{todonotes}

\newcommand{\cutparagraphup}{\vspace*{-0.12in}}

\icmltitlerunning{LLMs Can Automatically Engineer Features for Few-Shot Tabular Learning}

\begin{document}

\newcommand{\model}{\textsf{FeatLLM}}

\twocolumn[
\icmltitle{Large Language Models Can Automatically Engineer Features \linebreak for Few-Shot Tabular Learning}



\icmlsetsymbol{equal}{*}

\begin{icmlauthorlist}
\icmlauthor{Sungwon Han}{yyy}
\icmlauthor{Jinsung Yoon}{comp}
\icmlauthor{Sercan {\"O}. Arik}{comp}
\icmlauthor{Tomas Pfister}{comp}
\end{icmlauthorlist}

\icmlaffiliation{yyy}{Work done at Google as a research intern. School of Computing, Korea Advanced Institute of Science and Technology, Daejeon, Republic of Korea}
\icmlaffiliation{comp}{Google Cloud AI, Sunnyvale, California, USA}

\icmlcorrespondingauthor{Jinsung Yoon}{jinsungyoon@google.com}

\icmlkeywords{Machine Learning, ICML}

\vskip 0.3in
]



\printAffiliationsAndNotice{}  

\begin{abstract}
Large Language Models (LLMs), with their remarkable ability to tackle challenging and unseen reasoning problems, hold immense potential for tabular learning, that is vital for many real-world applications. 
In this paper, we propose a novel in-context learning framework, \model{}, which employs LLMs as feature engineers to produce an input data set that is optimally suited for tabular predictions.
The generated features are used to infer class likelihood with a simple downstream machine learning model, such as linear regression and yields high performance few-shot learning.
The proposed \model{} framework only uses this simple predictive model with the discovered features at inference time. 
Compared to existing LLM-based approaches, \model{} eliminates the need to send queries to the LLM for each sample at inference time.
Moreover, it merely requires API-level access to LLMs, and overcomes prompt size limitations.
As demonstrated across numerous tabular datasets from a wide range of domains, \model{} generates high-quality rules, significantly (10\% on average) outperforming alternatives such as TabLLM and STUNT.\looseness=-1
\end{abstract}

\section{Introduction}
In recent years, Large Language Models (LLMs) have shown impressive abilities for generating suitable responses for previously unseen tasks without requiring task-specific fine-tuning~\cite{creswell2022selection,imani2023mathprompter,taylor2022galactica}. Trained on extensive text corpora, the rich prior knowledge of LLMs can substitute the need for expert domain knowledge in many areas~\cite{Genomes2021,singhal2023large,wilcox2003role}, playing a crucial role in the automation of diverse applications, including dialogue analysis~\cite{finch2023leveraging} and program repair~\cite{fan2023automated}. Particularly, by incorporating a task description and few examples into prompts, LLMs can understand the context of a task and ``attend" to important features and examples~\cite{brown2020language,zhao2023survey}. This demonstrates the capability of LLMs to analyze data, highlighting their potential to perform as an expert feature engineer. \looseness=-1

Utilization of LLMs' prior knowledge and reasoning capabilities have been extended to the domain of tabular learning. For instance, knowledge such as related to the increased risk of certain diseases in older individuals can be beneficial at disease prediction tasks. In this context, recent works define task and feature descriptions in natural language, serialize the data, and then feed it into an LLM~\cite{dinh2022lift,wang2023anypredict}. This is often followed by either using in-context learning~\cite{nam2023semi} or parameter-efficient fine-tuning~\cite{dinh2022lift,hegselmann2023tabllm}. Prior knowledge provided by LLMs has proven effective, outperforming conventional tabular learning baselines in low-shot regimes~\cite{hegselmann2023tabllm}. \looseness=-1

Current LLM-based tabular learning methods have some limitations. For end-to-end predictions, at least one LLM inference per sample is required, making it computationally expensive. Most methods require fine-tuning the LLM for high accuracy, making them infeasible to the application to the LLMs that do not come with accessible training codebases or tuning APIs, a prominent concern given that recently-proposed top-performance LLMs only permit limited access via inference APIs~\cite{anil2023palm,openai2023gpt4}. In addition, most approaches are not suitable with lengthy prompts~\cite{liu2023lost} -- when the number of features in tabular data grows to the point where the serialized text length becomes long, the approach often turns infeasible or experiences a decline in performance. These constraints pose fundamental challenges to the real-world application of LLM-based tabular learning models. \looseness=-1

Our proposed approach goes beyond treating LLMs to yield end-to-end predictions. Instead, we assign them the role of feature engineering, with them proposing a more efficient way to utilize their prior knowledge. Specifically, we introduce a novel in-context learning framework, \model{}, which aims to understand the ``criteria" underlying LLM predictions. For instance, for the task of predicting a particular disease, the LLM can directly infer and generate rules that determine which feature conditions result in identifying the disease. The criteria (or rules) are used to create new features that can replace existing ones in solving tasks. This approach shifts the capability of the LLM to feature generation, yielding a simpler downstream model -- once the features are discovered, a complex machine learning model is no longer needed for inference, improving the inference latency.
By leveraging the in-context learning ability of LLMs, it's possible to extract features without the need for model training, by adding example demonstrations to the prompts. Furthermore, by employing techniques from ensemble methods~\cite{breiman2001random}, like feature bagging, we can mitigate the challenge of excessive prompt size. \looseness=-1

\model{} structures the input prompt for engineering new features into the following two sub-tasks: 
(1) understanding the problem and inferring the relationship between features and targets; and 
(2) based on the knowledge from the previous step and few-shot examples, deriving decisive rules to distinguish classes. This structured reasoning approach leads the LLMs to filter out uninformative features for rule extraction. The deduced rules are then applied to the data (parsed them as program code), and the data are transformed into a binary feature indicating whether or not the samples adheres to each rule. Finally, a model with low complexity is fitted to estimate class probabilities from these new features. The robustness is further improved by employing an ensemble approach, repeatedly generating features in conjunction with feature bagging. By appropriately adjusting the number of features and samples chosen in bagging, the challenge of excessive prompt size is mitigated. \model{} is free from training constraints and yields low inference cost, rendering the application of the LLM feasible for various real-world problems. \looseness=-1

We evaluate \model{} on 13 different tabular datasets in low-shot regimes, showing its strong and robust performance. We show that LLMs proficiently utilize both prior knowledge and the knowledge from the data, extracting high-quality features. Our framework outperforms contemporary few-shot learning baselines across various settings. The code is released via anonymized GitHub link at~\url{https://github.com/Sungwon-Han/FeatLLM}. \looseness=-1
\section{Related Work}

\begin{figure*}[t!]
\centerline{
      \includegraphics[width=1\linewidth]{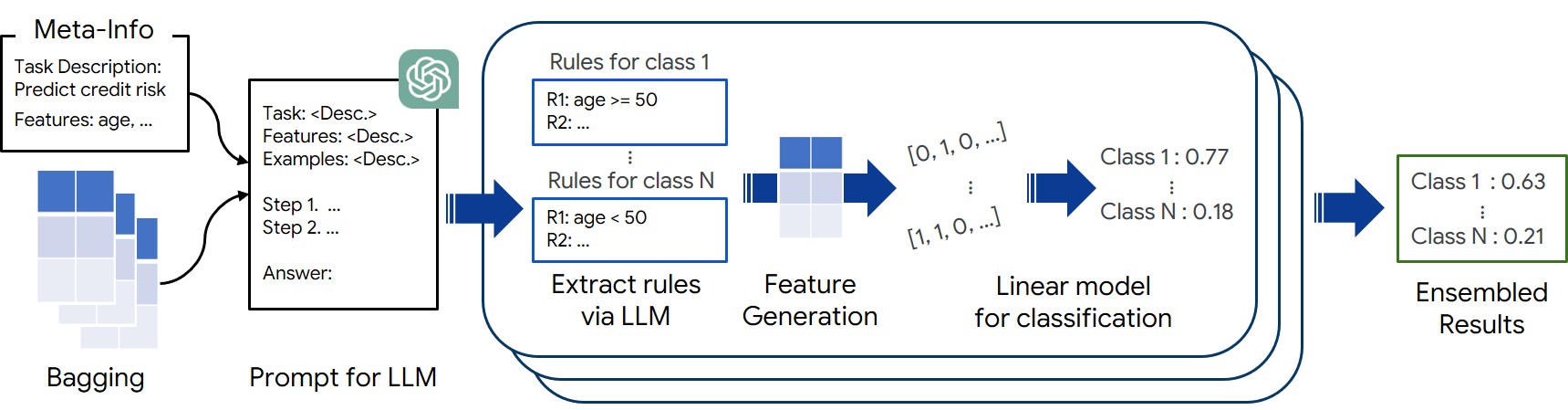}}
      \caption{Illustration of \model{}. \model{} extracts rules for each class, utilizing prior knowledge and few-shot examples. These rules are then parsed and applied to create binary features for data samples. A linear layer is trained on these binary features to estimate class likelihoods. This procedure is repeated multiple times for ensembling. \looseness=-1
      } 
\label{fig:main_model}
\end{figure*}

\subsection{Few-Shot Learning with Tabular Data}
The advancement of few-shot learning has enabled the extraction of generalizable representations from data, even with minimal labeling costs~\cite{chen2018closer}. Initially focusing on image data, few-shot learning has proven its strong generalizability across various data modalities including text, audio, and more recently, tabular data~\cite{majumder2022few,nam2023stunt,schick2021exploiting}. Learning from a small number of labeled samples presents the risk of learning spurious correlations~\cite{bartlett2021deep}. Consequently, recent efforts have employed additional sources of information or introduced domain-specific prior knowledge to impart appropriate inductive biases for model training. For example, utilizing large quantities of more readily-available unlabeled datasets with judicious augmentation for semi-supervised learning~\cite{ucar2021subtab,yoon2020vime}, or seeking good model initialization through task-agnostic unsupervised pre-training~\cite{somepalli2021saint}. Such unsupervised objectives can involve masking or corrupting part of the data and then using a reconstruction objective~\cite{arik2021tabnet,majmundar2022met} or employing data augmentation to apply contrastive learning objectives~\cite{bahri2022scarf,chen2020simple}. However, the heterogeneous nature of tabular data makes it challenging to find universally applicable augmentations, and these strategies have not been notably effective in few-shot settings~\cite{nam2023stunt}. \looseness=-1

Recent work have proposed training models on a broad range of tabular datasets beyond those related to the target task and then applying transfer learning to the target task~\cite{zhang2023generative,levin2022transfer,zhu2023xtab}. For instance, TransTab~\cite{wang2022transtab} trains transformer-based models learning of semantic relationships between columns through column names, not just table values. The knowledge extracted from various tables and columns has facilitated zero-shot or few-shot inference for new tabular tasks. Meanwhile, TabPFN~\cite{hollmann2022tabpfn} mixes different types of distributions to create synthetic data for pre-training a Bayesian neural network. After pre-training, inference is performed by inputting both training and test samples of the target task without further training. The prior knowledge gained from various datasets helped surpass the performance of traditional tree-based approaches and is proved to be beneficial in few-shot settings.

\subsection{Language-Interfaced Tabular Learning}
Another promising approach to leverage prior knowledge is the incorporation of large language models (LLMs)~\cite{anil2023palm,openai2023gpt4}. LLMs, trained on extensive and diverse text corpora, have demonstrated the ability to generalize to unseen tasks, proving their utility across various domains~\cite{brown2020language}. Recent studies have attempted to harness the rich prior knowledge encapsulated by LLMs for tabular learning. They typically propose serializing tabular data into text and feeding them into the input prompts of LLMs~\cite{dinh2022lift,hegselmann2023tabllm,wang2023anypredict}. By including tabular data, feature descriptions, and task descriptions in the input prompts, these can infer the relationships between tasks and features to perform inference. Research has evolved to either specifically train models using parameter-efficient tuning techniques like LoRA~\cite{hu2021lora} or IA3~\cite{liu2022few}, or employ in-context learning by adding few-shot example demonstrations to the prompts without training~\cite{dinh2022lift,hegselmann2023tabllm,nam2023semi,slack2023tablet}.

Despite the effectiveness of the aforementioned methods in improving few-shot tabular learning, the requirement to always pass inference through the LLM poses challenges for application in industrial domains. This study aims to depart from the conventional black-box approach of processing each sample's prediction in an end-to-end fashion via an LLM. Instead, it targets the LLM to directly infer the conditions or rules associated with each class. \model{} parses the generated rules into the program code to create new features, enabling predictions without the need to pass through the LLM and leveraging in-context learning to extract rules utilizing prior knowledge and information from few-shot samples. \looseness=-1
\section{Methods}
\model{} is designed to extract ``rules", the conditions associated with each answer class, from the few samples it inputs, as depicted in Figure~\ref{fig:main_model}. The extracted rules are parsed through an LLM, and then used to create new binary features of the given sample indicating whether or not each rule is satisfied. The set of new binary features are input to a dot product operation with non-negative trainable weights, to estimate the likelihood of each class. By training this model, we learn the importance of each feature and how it affects the class likelihood (Section~\ref{Sec:training}). This process is repeated multiple times with bagging and then combined via ensembling. Problem formulation and details of each stage are explained below. \looseness=-1

\cutparagraphup
\paragraph{Problem formulation.} Let's consider a tabular dataset with $N$ labeled samples $\mathcal{D} = \{(\mathbf{x}^i, \mathbf{y}^i)\}_{i=1}^{N}$. The dataset $\mathcal{D}$ consists of $d$ input features (i.e., each $\mathbf{x}^i$ is a $d$-dimensional vector), and comes with natural language feature names, denoted by $F = \{f_j\}_{j=1}^{d}$, along with their definitions provided separately. Assuming the classification setup, the label $\mathbf{y}^i$ is defined as one of the predefined set of classes. For the $k$-shot learning experiments, only $k$ ($<N$) labeled samples are randomly sampled to train the model. \looseness=-1

\begin{figure}[ht]
\begin{tcolorbox}[enhanced,attach boxed title to top center={yshift=-3mm,yshifttext=-1mm},
  colback=black!0!white, colframe=black!20!white, colbacktitle=black!10!white, coltitle=blue!20!black ]
\footnotesize{
You are an expert. Given the task description and the list of features and data examples, you are extracting conditions for each answer class to solve the task. \\

\textcolor{orange}{Task: $<$Task description$>$}

\textcolor{orange}{Features: $<$Feature descriptions$>$}

\textcolor{orange}{Examples: $<$Serialized training examples$>$} \\

\textcolor{blue}{Let's first understand the problem and solve the problem step by step.} \\

\textcolor{blue}{Step 1. Analyze the causal relationship or tendency between each feature and task description based on general knowledge and common sense within a short sentence.} \\

\textcolor{blue}{Step 2. Based on the above examples and Step 1's results, infer 10 different conditions per answer, following the format below. The condition should make sense, well match examples, and must match the format for [condition] according to value type.}\\

\textcolor{olive}{Format for Response: \\
10 different conditions for class [Class name]: \\
- [Condition]   \\
...\\}

\textcolor{olive}{Format for [Condition]: \\
For the categorical variable only, \\
- [Feature] is in [List of categories]\\
For the numerical variable only,\\
- [Feature] ($>$ or $>=$ or $<$ or $<=$) [Value]\\
- [Feature] is within range of [Value\_start, Value\_end]\\}

Answer:

Step 1. 
}
\end{tcolorbox}
\captionof{figure}{Prompt for rule extraction. Text in \textcolor{orange}{orange} provides basic information description; \textcolor{blue}{blue} text outlines reasoning instruction; and \textcolor{olive}{yellow} text details response instruction. \looseness=-1}
\label{fig:prompt_for_rule}            
\end{figure}

\subsection{Prompt Design for Extracting Rules}
\label{Sec:prompt_design}
To enable the LLM to extract rules based on a more accurate reasoning path, we guide the problem-solving process to mimic how an expert human might approach a tabular learning task. 
Our prompt comprises three main components as follows (shown in  Fig.~\ref{fig:prompt_for_rule} and Appendix~\ref{sec:example_rule}):

\cutparagraphup
\paragraph{Basic information description.} This part provides essential information for solving the problem (see orange text in Fig.~\ref{fig:prompt_for_rule}). It includes a description of the task to be solved with label information, descriptions of features, and example demonstrations. The task description is formulated as a question (e.g., ``Does this patient's myocardial infarction complications data indicate chronic heart failure? Yes or no?". More examples are provided in Appendix~\ref{sec:task_desc}). The feature description indicates its value type and optionally includes its definition. For categorical variables, examples of the possible categories can be provided. As example demonstrations, few training samples are serialized into text, along with their ground-truth labels. Serialization follows the format used in previous studies~\cite{dinh2022lift,hegselmann2023tabllm}: \looseness=-1
\begin{align}
&\text{Serialize}(\mathbf{x}^i,\mathbf{y}^i, F) = \nonumber \\ 
&``\ f_1\ \text{is}\ \mathbf{x}^i_1.\ ... \ f_d\  \text{is}\  \mathbf{x}^i_d.\ \ \text{Answer:}\  \mathbf{y}^i\ ”
\end{align}

\cutparagraphup
\paragraph{Reasoning instruction.} We aim to enhance the LLM's reasoning by providing guidance through reasoning instructions (see blue text in Fig.~\ref{fig:prompt_for_rule}). The reasoning instruction begins with an introductory sentence similar to the chain-of-thought approach~\cite{wei2022chain}.
Then, we divide the LLM's inference process into two steps. In the first step, the LLM is encouraged to infer the causal relationship or tendency between features and the task description. 
This is done without referring to example demonstrations, relying instead on general knowledge or common sense, allowing the LLM to fully extract prior knowledge about the problem. 
In the second step, the LLM uses example demonstrations and the information from the first step to deduce rules for each class. This two-step reasoning process prevents the model from identifying spurious correlations in irrelevant columns and can assist in focusing on more significant features. To prevent extracting too few or too many rules, potentially leading to either underfitting or creating an excessively large prompt, we have set a fixed number of rules (i.e., 10)\footnote{Analysis on the number of rules can be found in Section~\ref{sec:analysis}.} to be extracted for each class. Additionally, when formulating rules, we let the LLM refer to the type of each feature in the basic information description part to avoid type mismatches in rules. \looseness=-1

\cutparagraphup
\paragraph{Response instruction.} To facilitate easier parsing and use of the inferred rules, we guide the LLM on how to structure its responses through specific instructions (see yellow text in Fig.~\ref{fig:prompt_for_rule}). The prompt includes the guidelines on the required format for responses for each answer class (i.e., format for the response) and the structure each rule should follow (i.e., format for the rule). 
These instructions allow the LLM to select the appropriate format based on the feature type. Without additional guidance, the LLM can combine multiple rules using logical operators like AND or OR. The example outcome is provided in Appendix~\ref{sec:outcome-rule}. \looseness=-1

\subsection{Inferring Class Likelihood via Rules}
\label{Sec:training}

\begin{figure}[t!]
\begin{tcolorbox}[enhanced,attach boxed title to top center={yshift=-3mm,yshifttext=-1mm},
  colback=black!0!white, colframe=black!20!white, colbacktitle=black!10!white, coltitle=blue!20!black ]
\footnotesize{
Provide me a python code for function, given description below. \\

Function name: $<$Name$>$ \\
Input: Dataframe df\_input \\
Input Features: $<$Features$>$ \\
Output: Dataframe df\_output. Create a new dataframe df\_output. Each column in df\_output refers whether the selected column in df\_input follows the condition (1) or not (0). Be sure that the function code well matches with its feature type (i.e., numerical, categorical). \\

Conditions: \textcolor{blue}{$<$Conditions$>$} \\

Wrap only the function part with $<$start$>$ and $<$end$>$ tokens, and do not add any comments, descriptions, and package importing lines in the code.
}
\end{tcolorbox}
\captionof{figure}{Prompt for parsing rules. This prompt incorporates the rules generated in the previous stage, placed within the \textcolor{blue}{$<$Conditions$>$} section. \looseness=-1}
\label{fig:prompt_for_parsing}  
\end{figure}

\paragraph{Parsing rules for feature generation.} We utilize the rules generated in the previous stage to create new binary features. These features are created for each class, indicating whether the sample satisfies the rules associated with that class (i.e., either '0' or '1'). They can be used in predicting how likely a sample belongs to each class. However, since the rules generated by the LLM are based on natural language, designing a parsing strategy is required for automatic feature generation. Even though we constrain formatting of responses and rules during rule generation for easier parsing, the LLM's output can include noisy syntactic changes~\cite{kovavc2023large}. For instance, the LLM might alter expressions, like substituting ``$>$” or ``$<$” with phrases like ``greater than" or ``less than." This variability makes parsing more challenging. \looseness=-1

To address the challenges of parsing noisy text, instead of building complex program code, we leverage the LLM itself. The LLM is good at understanding the semantic meaning despite syntactic changes in text, and it has the ability to generate a simple code with instructions (given it's trained on code data corpus). To leverage these, we include the function name, input and output descriptions, and inferred rules in the prompt, then input them into the LLM. Figures~\ref{fig:prompt_for_parsing} and ~\ref{sec:example_parsing}-\ref{sec:example_parse_outcome} in Appendix show the prompt examples and the corresponding outcomes. The generated code is executed using Python’s \texttt{exec()} function to perform data conversion. \looseness=-1


\cutparagraphup
\paragraph{Inferring class likelihood.} 
Given new binary features from rules for each class, a simple method to measure the class likelihood of the sample is to count how many rules of each class it satisfies (i.e., the sum of the binary feature per class). However, not all rules and features carry the same importance, necessitating learning their significance from training samples. 
We aim to learn this importance using a conventional machine learning (ML) model, applied to each class's binary feature. 
For example, we can consider a linear model without bias.
Let us denote the generated binary feature from sample $\mathbf{x}^i$ for class $k$ as $\mathbf{z}_k^i \in \{0, 1\}^R$, where $R$ is the number of rules per each class and $c$ is the number of classes. The probability for each class $\mathbf{p}^i$ is computed as follows: \looseness=-1
\begin{align}
\text{logit}_k^i &= \max(\mathbf{w}_k, 0) \cdot \mathbf{z}_k^i, \nonumber \\
\mathbf{p}^i &= \text{Softmax}({[\text{logit}_{1}^i, ..., \text{logit}_{c}^i]}). \label{eq:infer_prob}
\end{align}
Here, $\mathbf{w}_k$ represent the trainable weight in the linear model and they signify the importance of each feature. By projecting these weights into a positive subspace, we can ensure that the features generated by the LLM do not reduce the likelihood of a class during learning.
The logit vector is then converted into class probabilities through the softmax function. We optimize the cross-entropy loss to learn the weight $\mathbf{w}_k$ using a few-shot training samples. \looseness=-1



\cutparagraphup
\paragraph{Ensembling with bagging.}
Finally, we repeatedly execute the entire process to create multiple inference models to make the final prediction via ensemble. Given the collection of trained weights from $T$ trials, denoted as $\{\mathbf{w}[t]\}_{t=1}^T$, the prediction is made by averaging the probability for each class $\mathbf{p}^i$ computed with each weight $\mathbf{w}[t]$ as in Eq.~\ref{eq:infer_prob}.

To introduce randomness in rule generation for ensemble, a sufficiently high temperature is set for LLM inference. We also randomize the order of few-shot demonstrations based on the observation that the order can impact the query outcome of LLM~\cite{lu2022fantastically}\footnote{Analysis on the rule diversity can be found in Appendix~\ref{sec:diversity}}. To utilize the diversity in prediction for more robust predictions, we adopt bagging to select a subset of features or instances for each trial, which gets particularly useful for larger number of training samples or features. \looseness=-1
The proposed ensemble approach offers two main advantages. First, even if the LLM generates incorrect rules in some trials, the correct rules from other trials can compensate, enhancing the overall robustness of the framework. This ties into the LLM's self-consistency~\cite{wang2022self}, as rules commonly inferred across multiple trials are more likely to be accurate.  
Second, the ensemble method addresses the limitation of LLM's prompt size. When the tabular data exceed the allowable input prompt size, bagging helps reduce it, ensuring the model's robust performance. While a single rule set might not capture the relationships of all features, the ensemble of various weak learners created across multiple trials can nullify the impact of incomplete feature/instance coverage in individual trials.\looseness=-1

\begin{table*}[t!]
\setlength{\tabcolsep}{3.5pt}
\centering
\caption{Evaluation results, showing AUC across 11 datasets with attributes fewer than 100, are presented. Best performances are bolded, and our framework's performances, when second-best, are underlined. Results with more training shots (e.g., 32, 64) and an additional baseline (e.g., RandomForest) are provided in Appendix~\ref{sec:full}.}
\scalebox{0.81}{
\begin{tabular}{@{}l|c|cccccccc|c@{}}
\toprule
Data & Shot & LogReg & XGBoost & SCARF & TabPFN & STUNT & In-context & TABLET & TabLLM & Ours \\ \midrule
Adult & 4 & 72.10$\pm$12.30 & 50.00$\pm$0.00 & 58.34$\pm$15.42 & 60.89$\pm$23.28 & 67.43$\pm$29.61 & 77.51$\pm$5.24 & 75.29$\pm$12.24 & 83.57$\pm$2.69 & \textbf{86.68$\pm$0.86} \\
 & 8 & 76.02$\pm$3.37 & 59.19$\pm$6.92 & 72.42$\pm$8.95 & 70.42$\pm$9.96 & 82.16$\pm$6.93 & 79.30$\pm$2.89 & 77.56$\pm$7.56 & 83.52$\pm$4.30 & \textbf{87.89$\pm$0.06} \\
 & 16 & 75.20$\pm$5.10 & 60.68$\pm$13.92 & 75.63$\pm$9.56 & 70.34$\pm$9.96 & 80.57$\pm$10.93 & 79.50$\pm$4.57 & 79.74$\pm$5.64 & 83.23$\pm$2.45 & \textbf{87.54$\pm$0.50} \\ \midrule
Bank & 4 & 63.70$\pm$3.87 & 50.00$\pm$0.00 & 58.53$\pm$5.49 & 63.19$\pm$11.60 & 56.34$\pm$12.82 & 61.38$\pm$1.30 & 58.11$\pm$6.29 & 62.51$\pm$8.95 & \textbf{70.45$\pm$3.69} \\
 & 8 & 72.52$\pm$3.21 & 58.78$\pm$10.54 & 55.28$\pm$11.88 & 62.81$\pm$7.84 & 63.01$\pm$8.78 & 69.57$\pm$13.35 & 69.08$\pm$6.00 & 63.19$\pm$5.79 & \textbf{75.85$\pm$6.66} \\
 & 16 & 77.51$\pm$3.09 & 70.34$\pm$5.86 & 65.81$\pm$1.79 & 73.79$\pm$2.21 & 69.85$\pm$0.95 & 69.76$\pm$8.55 & 69.40$\pm$11.28 & 63.73$\pm$6.43 & \textbf{78.41$\pm$1.08} \\ \midrule
Blood & 4 & 56.79$\pm$26.02 & 50.00$\pm$0.00 & 56.22$\pm$21.00 & 58.72$\pm$19.16 & 48.57$\pm$6.04 & 56.30$\pm$12.43 & 56.45$\pm$15.45 & 55.87$\pm$13.49 & \textbf{68.34$\pm$7.48} \\
 & 8 & 68.51$\pm$5.16 & 59.97$\pm$1.36 & 65.77$\pm$5.00 & 66.30$\pm$10.01 & 60.00$\pm$4.84 & 58.99$\pm$10.12 & 56.37$\pm$11.56 & 66.01$\pm$9.25 & \textbf{70.37$\pm$3.23} \\
 & 16 & 68.30$\pm$6.16 & 63.28$\pm$7.62 & 66.27$\pm$5.04 & 64.14$\pm$6.80 & 54.76$\pm$4.53 & 56.59$\pm$5.21 & 60.62$\pm$4.13 & 65.14$\pm$7.55 & \textbf{70.07$\pm$5.19} \\ \midrule
Car & 4 & 62.38$\pm$4.13 & 50.00$\pm$0.00 & 62.52$\pm$3.80 & 58.14$\pm$4.15 & 61.32$\pm$3.83 & 62.47$\pm$2.47 & 60.21$\pm$4.81 & \textbf{85.82$\pm$3.65} & {\ul 72.69$\pm$1.52} \\
 & 8 & 72.05$\pm$1.20 & 64.00$\pm$3.57 & 72.23$\pm$2.59 & 63.95$\pm$4.35 & 67.86$\pm$0.49 & 67.57$\pm$3.44 & 65.53$\pm$8.00 & \textbf{87.43$\pm$2.56} & {\ul 73.26$\pm$1.46} \\
 & 16 & 82.42$\pm$4.13 & 72.26$\pm$4.43 & 75.77$\pm$2.71 & 71.35$\pm$5.33 & 75.56$\pm$2.88 & 76.94$\pm$3.04 & 74.02$\pm$1.01 & \textbf{88.65$\pm$2.63} & 79.43$\pm$1.24 \\ \midrule
Credit-g & 4 & 52.68$\pm$4.46 & 50.00$\pm$0.00 & 48.92$\pm$4.60 & 54.00$\pm$7.34 & 48.80$\pm$6.76 & 52.99$\pm$4.08 & 54.33$\pm$6.54 & 51.90$\pm$9.40 & \textbf{55.94$\pm$1.10} \\
 & 8 & 55.52$\pm$8.88 & 52.22$\pm$4.90 & 55.26$\pm$3.92 & 52.58$\pm$11.27 & 54.50$\pm$8.25 & 52.43$\pm$4.36 & 52.90$\pm$5.79 & 56.42$\pm$12.89 & \textbf{57.42$\pm$3.10} \\
 & 16 & 58.26$\pm$5.17 & 56.23$\pm$4.37 & 59.22$\pm$11.38 & 58.91$\pm$8.04 & 57.63$\pm$7.58 & 55.29$\pm$4.80 & 51.65$\pm$4.02 & \textbf{60.38$\pm$14.03} & 56.60$\pm$2.22 \\ \midrule
Diabetes & 4 & 57.09$\pm$18.84 & 50.00$\pm$0.00 & 62.35$\pm$7.48 & 56.28$\pm$13.01 & 64.22$\pm$6.78 & 71.71$\pm$5.31 & 63.96$\pm$3.32 & 70.42$\pm$3.69 & \textbf{80.28$\pm$0.75} \\
 & 8 & 65.52$\pm$13.18 & 50.86$\pm$22.03 & 64.69$\pm$13.33 & 69.08$\pm$9.68 & 67.39$\pm$12.92 & 72.21$\pm$2.07 & 65.47$\pm$3.95 & 64.30$\pm$5.88 & \textbf{79.38$\pm$1.66} \\
 & 16 & 73.44$\pm$0.52 & 65.69$\pm$6.54 & 71.86$\pm$3.16 & 73.69$\pm$3.21 & 73.79$\pm$6.48 & 71.64$\pm$5.05 & 66.71$\pm$0.76 & 67.34$\pm$2.79 & \textbf{80.15$\pm$1.35} \\ \midrule
Heart & 4 & 70.54$\pm$3.83 & 50.00$\pm$0.00 & 59.38$\pm$3.42 & 67.33$\pm$15.29 & \textbf{88.27$\pm$3.32} & 60.76$\pm$4.00 & 68.19$\pm$11.17 & 59.74$\pm$4.49 & {\ul 75.66$\pm$4.59} \\
 & 8 & 78.12$\pm$10.59 & 55.88$\pm$3.98 & 74.35$\pm$6.93 & 77.89$\pm$2.34 & \textbf{88.78$\pm$2.38} & 65.46$\pm$3.77 & 69.85$\pm$10.82 & 70.14$\pm$7.91 & {\ul 79.46$\pm$2.16} \\
 & 16 & 83.02$\pm$3.70 & 78.62$\pm$7.14 & 83.66$\pm$5.91 & 81.45$\pm$5.05 & \textbf{89.13$\pm$2.10} & 67.00$\pm$7.83 & 68.39$\pm$11.73 & 81.72$\pm$3.92 & {\ul 83.71$\pm$1.88} \\ \midrule
Cultivars & 4 & 53.45$\pm$10.79 & 50.00$\pm$0.00 & 46.99$\pm$6.33 & 49.80$\pm$15.90 & \textbf{57.10$\pm$8.66} & 51.38$\pm$2.48 & 54.28$\pm$3.73 & 54.39$\pm$5.61 & {\ul 55.63$\pm$5.24} \\
 & 8 & 56.22$\pm$11.87 & 52.60$\pm$6.31 & 51.76$\pm$9.99 & 54.72$\pm$9.35 & \textbf{57.26$\pm$9.52} & 51.68$\pm$4.43 & 51.48$\pm$3.85 & 52.86$\pm$6.13 & {\ul 56.97$\pm$5.08} \\
 & 16 & \textbf{60.35$\pm$4.23} & 56.87$\pm$2.50 & 57.06$\pm$9.27 & 54.92$\pm$8.32 & 60.09$\pm$7.64 & 54.31$\pm$6.12 & 57.44$\pm$3.53 & 56.97$\pm$2.22 & 57.19$\pm$5.30 \\ \midrule
NHANES & 4 & 91.96$\pm$7.02 & 50.00$\pm$0.00 & 51.58$\pm$4.66 & 80.74$\pm$3.89 & 69.32$\pm$19.59 & 91.84$\pm$3.79 & 93.54$\pm$4.20 & \textbf{99.49$\pm$0.23} & 92.20$\pm$1.71 \\
 & 8 & 92.38$\pm$8.21 & 92.92$\pm$4.56 & 55.07$\pm$4.79 & 85.10$\pm$5.88 & 68.56$\pm$18.35 & 86.67$\pm$5.49 & 94.25$\pm$3.35 & \textbf{100.00$\pm$0.00} & 93.29$\pm$7.01 \\
 & 16 & 94.12$\pm$3.88 & 94.52$\pm$6.28 & 89.78$\pm$6.77 & 95.54$\pm$0.82 & 68.62$\pm$19.81 & 93.33$\pm$4.47 & 95.02$\pm$1.57 & \textbf{100.00$\pm$0.00} & {\ul 95.64$\pm$4.67} \\ \midrule
Sequence & 4 & 66.80$\pm$4.98 & 50.00$\pm$0.00 & 67.08$\pm$2.75 & 71.93$\pm$4.17 & 60.12$\pm$6.51 & 93.01$\pm$1.84 & 91.00$\pm$4.25 & 75.38$\pm$5.21 & \textbf{98.60$\pm$0.66} \\
-type & 8 & 72.55$\pm$4.37 & 61.01$\pm$10.96 & 60.53$\pm$6.91 & 79.55$\pm$6.71 & 60.58$\pm$9.58 & 94.29$\pm$1.64 & 90.18$\pm$3.73 & 79.91$\pm$2.13 & \textbf{98.70$\pm$1.44} \\
 & 16 & 83.10$\pm$6.86 & 80.31$\pm$5.79 & 73.14$\pm$5.29 & 89.59$\pm$2.33 & 65.12$\pm$5.28 & 93.46$\pm$0.72 & 92.09$\pm$2.90 & 86.97$\pm$1.97 & \textbf{99.29$\pm$0.78} \\ \midrule
Solution & 4 & 72.71$\pm$1.96 & 50.00$\pm$0.00 & 68.48$\pm$14.97 & 71.19$\pm$3.90 & 64.52$\pm$8.41 & 73.53$\pm$1.94 & 80.26$\pm$5.11 & 73.75$\pm$3.90 & \textbf{100.00$\pm$0.00} \\
-mix & 8 & 82.91$\pm$7.74 & 58.31$\pm$19.17 & 56.67$\pm$7.64 & 82.42$\pm$3.72 & 72.04$\pm$10.34 & 76.34$\pm$4.26 & 83.19$\pm$3.70 & 76.57$\pm$4.80 & \textbf{99.81$\pm$0.08} \\
 & 16 & 83.87$\pm$1.90 & 68.09$\pm$1.76 & 70.37$\pm$13.96 & 84.39$\pm$0.83 & 71.82$\pm$8.79 & 82.15$\pm$5.01 & 88.10$\pm$4.72 & 84.28$\pm$5.15 & \textbf{99.67$\pm$0.13} \\ \midrule
Average & 4 & 65.47 & 50.00 & 58.22 & 62.93 & 62.36 & 68.44 & 68.69 & 70.26 & \textbf{77.86} \\
 & 8 & 72.03 & 60.52 & 62.18 & 69.53 & 67.47 & 70.41 & 70.53 & 72.76 & \textbf{79.31} \\
 & 16 & 76.33 & 69.72 & 71.69 & 74.37 & 69.72 & 72.72 & 73.02 & 76.22 & \textbf{80.70} \\ \bottomrule
\end{tabular}}
\label{tab:main1}
\end{table*}

\section{Experiments}
We evaluate \model{} over multiple tabular datasets, and conduct various analyses to investigate how our framework works. Due to space constraints, additional experiments, such as varying LLMs, qualitative analysis, and others, are described in Appendix. \looseness=-1

\subsection{Performance Evaluation}
\paragraph{Datasets.}
Our experiment utilizes 13 datasets for binary or multi-class classification tasks: (1) Adult~\cite{asuncion2007uci} for predicting whether an individual earns over \$50,000 annually; (2) Bank~\cite{moro2014data} for predicting whether a customer will subscribe to a term deposit; (3) Blood~\cite{yeh2009knowledge} for predicting whether donors will return for subsequent donations; (4) Car~\cite{kadra2021well} for predicting the quality of a car; (5) Communities~\cite{misc_communities_and_crime_183} for predicting the crime rate in a specified region; (6) Credit-g~\cite{kadra2021well} for predicting whether an individual poses a good or bad credit risk; (7) Diabetes\footnote{\url{kaggle.com/datasets/uciml/pima-indians-diabetes-database}} for predicting the presence of diabetes in an individual; (8) Heart\footnote{\url{kaggle.com/datasets/fedesoriano/heart-failure-prediction}} for predicting the occurrence of coronary artery disease in an individual; and (9) Myocardial~\cite{misc_myocardial_infarction_complications_579} for predicting whether an individual suffers from chronic heart failure. 

We also include more recent datasets and synthetic datasets which have been less frequently discussed and were not included in the pre-training process of LLMs: (10) Cultivars for predicting how high the grain yield of this soybean cultivar will be\footnote{\url{archive.ics.uci.edu/dataset/913}}; (11) NHANES for predicting the given person's age group from the record\footnote{\url{archive.ics.uci.edu/dataset/887}}; (12) Sequence-type for classifying a given sequence as one of four types: arithmetic, geometric, Fibonacci, or Collatz; (13) Solution-mix for predicting whether the percentage concentration of a mixture from four solutions exceeds 0.5. Former two datasets were recently donated to the UCI Machine Learning Repository after September 2023, ensuring they were not part of the pre-training for the LLM API (gpt-3.5-turbo-0613) used by our model. The latter two are synthetic datasets which ensures they could not be memorized by the LLM API and proceeded with the evaluation.

These datasets vary in size and complexity, as detailed in Table~\ref{Tab:basic-desc}.
Each dataset provides a clear name and description for each attribute. 
Datasets like `Communities' and `Myocardial' include over 100 attributes, posing challenges due to limited context window size in LLM. \looseness=-1

\begin{table}[ht!]
\centering
\caption{Basic information of each dataset used in our experiments. The numbers in parentheses indicate the count of categorical and numerical attributes, respectively.}
\scalebox{0.85}{
\begin{tabular}{@{}l|ccc@{}}
\toprule
Data     & \multicolumn{1}{l}{\# of samples} & \multicolumn{1}{l}{\# of features} & \multicolumn{1}{c}{Label ratio (\%)} \\ \midrule
Adult    & 48842                             & 14 (7/7)                                & 76:24                                       \\
Bank     & 45211                             & 16 (8/8)                           & 88:12                                        \\
Blood    & 748                               & 4 (0/4)                                  & 76:24                                         \\
Car      & 1728                              & 6 (5/1)                            & 70:22:4:4                               \\
Communities   & 1994                          & 103 (1/102)                              & 34:33:33                               \\
Credit-g & 1000                              & 20 (12/8)                                 & 70:30                                        \\
Diabetes & 768                               & 8 (0/8)                                  & 65:35                                        \\
Heart    & 918                               & 11 (4/7)                                 & 45:55                                        \\
Myocardial    & 1700                          & 111 (94/17)                                & 22:78                                        \\
{Cultivars}     & {320}                          & {10 (3/7)}                                & {50:50}                                        \\
{NHANES}     & {6287}                          & {8 (1/7)}                                & {84:16}                                        \\
{Sequence-type}    & {250}                          & {5 (0/5)}                                & {40:40:10:10}                                        \\
{Solution-mix}     & {300}                          & {8 (0/8)}                                & {52:48}                                        \\
\bottomrule
\end{tabular}
}
\label{Tab:basic-desc}
\end{table}

\cutparagraphup
\paragraph{Baselines.}
We conduct comparisons with nine baselines. The first three baselines are conventional tabular learning approaches, including (1) Logistic regression (LogReg), (2) XGBoost~\cite{chen2016xgboost}, and (3) RandomForest~\cite{ho1995random}. The next two baselines assume the use of unlabeled datasets, including (4) SCARF~\cite{bahri2022scarf} and (5) STUNT~\cite{nam2023stunt}. Note that obtaining a large quantity of unlabeled dataset may be infeasible in real-world applications. Another recent baseline is (6) TabPFN~\cite{hollmann2022tabpfn}, which generates a synthetic dataset with diverse distribution to pretrain the model. The final three baselines utilize LLM, including (7) In-context learning (In-context)~\cite{wei2022emergent}, (8) TABLET~\cite{slack2023tablet}, and (9) TabLLM~\cite{hegselmann2023tabllm}. In-context learning directly embeds few-shot training samples into the input prompt without tuning. TABLET further incorporates additional hints into the prompt through an external classifier via rule-sets and prototypes to enhance the quality of inference. TabLLM employs parameter-efficient tuning technique like IA3~\cite{liu2022few} to train the LLM with few-shot samples. 

\looseness=-1

\cutparagraphup
\paragraph{Implementation details.}
\model{} uses GPT-3.5 as its LLM backbone, yet is designed to be agnostic to the choice of LLMs (see Appendix~\ref{sec:backbones} for results with different backbones). 
The temperature for the LLM inference is set to 0.5 and the top-p value is set to the default value of 1 in API. We set the number of ensembles and the number of rules for extracting to 20 and 10 respectively. Details on hyper-parameter impacts are in Figure~\ref{fig:hyperparameter2} and Appendix~\ref{sec:hyper-parameter}.
We use the Adam optimizer of a learning rate 0.01 for the linear model, training for 200 epochs. We employ $k$-fold cross-validation for optimal epoch selection. In replicating baselines, GPT-3.5 is utilized for in-context learning, whereas T0~\cite{sanh2022multitask} is employed in the TabLLM, following the original setting. Note that TabLLM restricts the use of LLM to only those with publicly available checkpoints; therefore, GPT-3.5 is not accessible in this context. Regarding traditional machine learning methods, such as LogReg, XGBoost, and RandomForest, we determined their parameters using Grid-search combined with $k$-fold cross validation. The value of $k$ is set either as 2 or 4, guaranteeing that the training set includes at least one example of each class. For more detailed information on implementations, please refer to Appendix~\ref{sec:baselines}. \looseness=-1

\begin{table}[t!]
\centering
\caption{Evaluation results with AUC across two datasets with over 100 attributes. TabPFN and two LLM-based baselines are excluded in the table as they failed to make inference over the dataset with a large number of features.}
\scalebox{0.85}{
\begin{tabular}{@{}l|ccc@{}}
\toprule
\multirow{2}{*}{Communities} & \multicolumn{3}{c}{Shots} \\ \cmidrule(l){2-4} 
 & 4 & 8 & 16 \\ \midrule
LogReg & 67.45$\pm$13.26 & 73.73$\pm$5.45 & 72.55$\pm$4.83 \\
XGBoost & 53.94$\pm$4.19 & 66.65$\pm$4.50 & 68.01$\pm$1.97 \\
RandomForest & 66.09$\pm$10.52 & 71.16$\pm$4.61 & 71.66$\pm$4.81 \\
SCARF & 66.18$\pm$9.13 & 72.69$\pm$3.79 & 73.09$\pm$2.84 \\
STUNT & 66.87$\pm$14.10 & 76.36$\pm$4.55 & \textbf{77.29$\pm$2.56} \\
\midrule
\model{} & \textbf{75.39$\pm$5.05} & \textbf{76.59$\pm$1.25} & {\ul 76.25$\pm$0.64} \\ 
\midrule
\midrule
\multirow{2}{*}{Myocardial} & \multicolumn{3}{c}{Shots} \\ \cmidrule(l){2-4} 
 & 4 & 8 & 16 \\ \midrule
LogReg & 51.25$\pm$3.85 & 55.34$\pm$1.11 & 60.00$\pm$5.16 \\
XGBoost & 50.00$\pm$0.00 & 55.63$\pm$2.92 & 56.55$\pm$12.22 \\
RandomForest & 51.91$\pm$4.49 & 52.77$\pm$5.83 & 54.16$\pm$4.53 \\
SCARF & 47.70$\pm$4.10 & 49.37$\pm$3.41 & 54.31$\pm$1.42 \\
STUNT & 52.77$\pm$2.01 & 55.40$\pm$4.41 & \textbf{61.22$\pm$3.45} \\
\midrule
\model{} & \textbf{52.87$\pm$3.44} & \textbf{56.22$\pm$1.64} & 55.32$\pm$9.15 \\ \bottomrule
\end{tabular}}
\label{tab:main2}
\end{table}

\begin{figure}[t!]
\centerline{
      \includegraphics[width=1\linewidth]{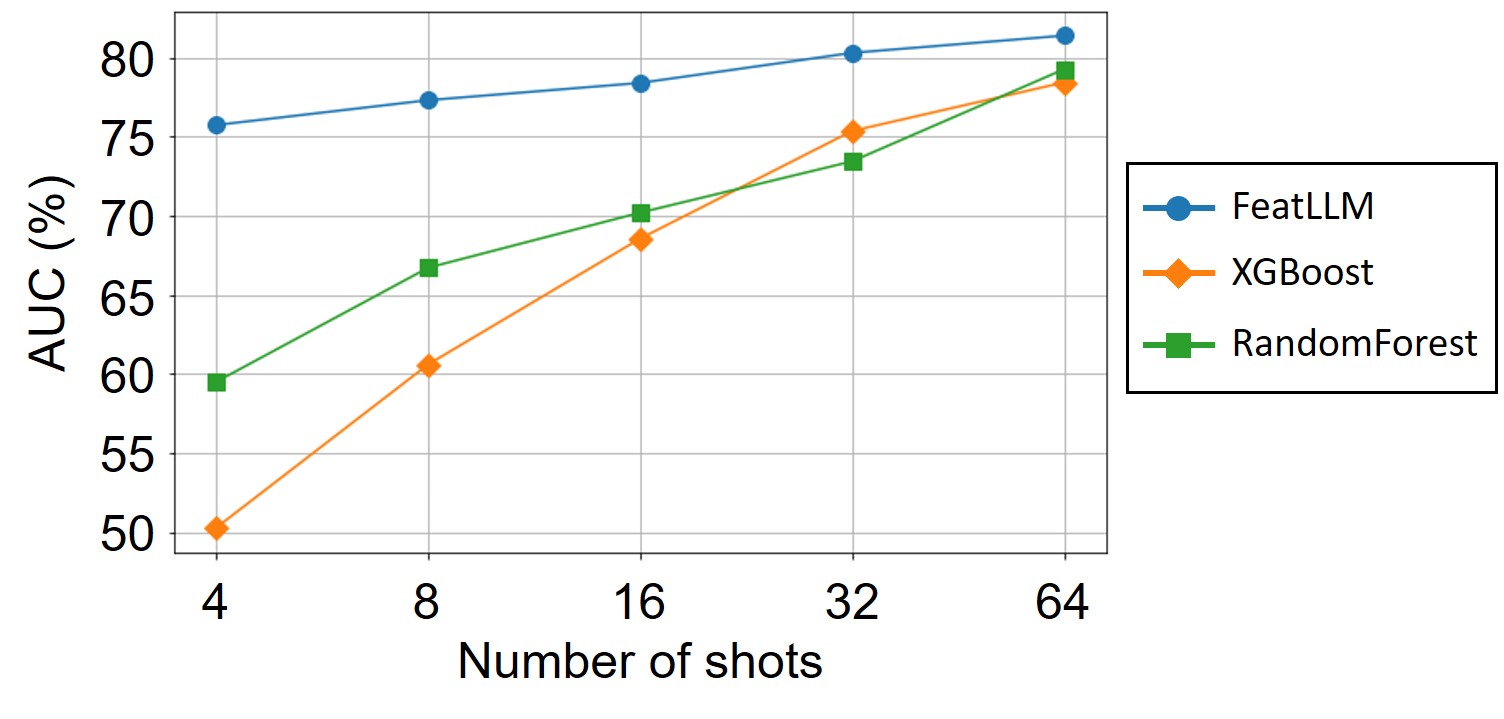}}
      \caption{Performance comparison summaries among conventional tabular baselines and \model{}. Averaged AUC over all datasets across the number of shots are reported. \looseness=-1
      } 
\label{fig:compares}
\end{figure}

\cutparagraphup
\paragraph{Results.}
In Tables~\ref{tab:main1} and \ref{tab:main2}, performance comparisons across a range of datasets are presented. We repeat experiments three times with different randomness on the choice of training/test sets, and present the average and standard deviation of the AUC values. \model{} consistently ranks as the top performer or secures the second place when compared with other baseline models. Our framework, when utilizing just 4 shots, achieves comparable performance to conventional tabular baselines that require 32 shots (see Fig.~\ref{fig:compares}). Notably, while STUNT and TabPFN demonstrated significant performance improvements on certain datasets, their overall enhancement over conventional machine learning models was not substantial. 
Moreover, models utilizing LLM, such as in-context learning, TABLET, or TabLLM, failed to infer on tabular datasets with a large number of features. Our framework, employing both bagging and ensemble techniques, proved to be effective even on datasets with numerous features, exhibiting satisfactory performance. It is important to note that our bagging and ensemble concept could be adapted to existing LLM-based approaches. However, such an adaptation would double the number of per-sample inference, leading to a substantial increase in computational complexity and thus becoming impractical. \looseness=-1

\subsection{Analysis \& Discussion}
\label{sec:analysis}
\paragraph{Ablation study.}
We analyze the contributions of major components to the performance by performing ablation studies, focusing on: (1) \textsf{-Tuning}: omitting the weight tuning process in the linear model, summing up the binary feature to compute the logit instead; (2) \textsf{-Ensemble}: omitting the ensemble process; (3) \textsf{-Description}: omitting the feature description, and (4) \textsf{-Reasoning}: omitting the Step-1 process in reasoning instruction part when generating rules.

Table~\ref{tab:ablation} summarizes the change in the AUC from ablations averaged across all datasets. In all cases, modifying the ablated component results in performance degradation. Interestingly, the benefit of weight tuning becomes higher when the number of shots increases. This indicates that when there is a large amount of data, accurate estimation of the importance of rules becomes feasible, making the effect of tuning significant. On the other hand, the effect of feature descriptions and reasoning instructions are high when the number of shot is small. This suggests that the efficient utilization of the prior knowledge of LLM becomes crucial for performance improvements. Lastly, the proposed ensemble approach consistently improves the performance in all settings. \looseness=-1

\begin{table}[h!]
\centering
\setlength{\tabcolsep}{4.5pt}
\caption{Ablation study summaries. Averaged changes of AUC with standard error after altering or omitting modules over 13 datasets with three different trials are reported.}
\scalebox{0.79}{
\begin{tabular}{@{}c|c|cccc@{}}
\toprule
Shot & \model{} & -Tuning & -Ensemble & -Description & -Reasoning \\ \midrule
4 & 75.7 & -1.41$\pm$1.00 & -5.39$\pm$0.81 & -1.76$\pm$1.06 & -5.03$\pm$1.96 \\
8 & 77.3 & -2.72$\pm$0.93 & -6.96$\pm$1.40 & -1.20$\pm$0.33 & -3.55$\pm$0.81 \\
16 & 78.4 & -2.57$\pm$0.73 & -6.65$\pm$1.18 & -0.26$\pm$0.31 & -1.50$\pm$0.87 \\
32 & 80.3 & -5.75$\pm$1.19 & -7.38$\pm$1.34 & -0.29$\pm$0.58 & -2.42$\pm$1.15 \\
64 & 81.4 & -4.88$\pm$1.40 & -6.09$\pm$0.96 & -0.70$\pm$0.54 & -1.71$\pm$0.47 \\
Avg & 78.6 & -3.47$\pm$0.51 & -6.49$\pm$0.51 & -0.84$\pm$0.28 & -2.84$\pm$0.53 \\ \bottomrule
\end{tabular}}
\label{tab:ablation}
\end{table}

\cutparagraphup
\paragraph{Training \& inference time comparison.}
We compare the training and inference time between different methods. The experiments are performed on the Adult dataset, including a training set of 16 shots. One A100 GPU is used as the default, except for TabLLM which uses four A100 GPUs for model parallelism. Table~\ref{tab:cost} provides the runtime comparisons. Notably, \model{} shows a relatively low inference time, comparable to that of conventional methods. In contrast, other LLM-based approaches, such as In-context learning, TABLET, and TabLLM, exhibit much higher inference time. In terms of training time, \model{} is comparable with other LLM-based methods -- it requires only 30 API queries, which does not impose a significant burden on the budget and even can be parallelized. \looseness=-1
\begin{table}[t!]
\setlength{\tabcolsep}{4pt}
\centering
\caption{Comparison of training (with 16 samples) and inference (over one sample) time on Adult.}
\scalebox{0.87}{
\begin{tabular}{@{}l|cc@{}}
\toprule
Model & Training (in seconds)  & Inference (in milliseconds)  \\ \midrule
LogReg & 0.721 & 0.001 \\
XGBoost & 28.512 & 0.006 \\
RandomForest & 1.343 & 0.001 \\
SCARF & 426.859 & 0.002 \\
TabPFN & 0.440 & 1.149 \\
STUNT & 642.796 & 0.006 \\
In-context$\dagger$ & N/A & 463.000 \\
TABLET$\dagger$ & 0.813 & 523.254 \\
TabLLM & 251.242 & 335.127 \\
\midrule
\model{}$\dagger$ & 860.094 & 0.006 \\ \bottomrule
\end{tabular}
} 
\begin{flushleft}
$\dagger$ \small{These models employ API queries, where the runtime is subject to the API's status at the time of use. \looseness=-1}
\end{flushleft}
\label{tab:cost}
\end{table}

\cutparagraphup
\paragraph{Quality of features generated by \model{}.}
We conduct a simple experiment to investigate the informativeness of rule-based features generated by \model{} in solving practical tasks. 
Specifically, we measure the average AUROC between the newly generated features and the target labels, and then compute the ratio of useful features with an AUROC higher than 0.5 for each dataset. 
Table~\ref{tab:quality} below summarizes the results.
It confirms that the majority of the created rules are relevant to the target attribute and provide informative insights for addressing the target task (refer to the column ``Before tuning"). 
Moreover, during the process of tuning the linear model with data, rules that have little relevance to the data (i.e., rules with learned weights below the threshold) are automatically filtered out (refer to the column ``After tuning").
Note that the threshold was set at 0.1, taking into consideration the overall histogram of weights. \looseness=-1

\begin{table}[h!]
\centering
\caption{Evaluation of feature quality before and after linear model tuning. Ratio of newly-generated features with an AUROC higher than 0.5 is reported across datasets.}
\scalebox{0.83}{
\begin{tabular}{@{}l|cc@{}}
\toprule
\multirow{2}{*}{Method} & \multicolumn{2}{c}{Ratio of useful rules} \\ \cmidrule{2-3} 
& Before tuning & After tuning \\ \midrule
Adult & 0.971 & 0.980 \\
Bank & 0.627 & 0.958 \\
Blood & 0.889 & 0.975 \\
Car & 0.712 & 0.849 \\
Communities & 0.767 & 0.889 \\
Credit-g & 0.643 & 0.802 \\
Diabetes & 0.974 & 0.972 \\
Heart & 0.800 & 0.924 \\
Myocardial & 0.690 & 0.744 \\
Cultivars & 0.775 & 0.927 \\
NHANES & 0.600 & 0.832 \\
Sequence-type & 0.896 & 0.985 \\
Solution-mix & 0.917 & 1.000 \\ \midrule
Average & 0.789 & 0.911 \\ \bottomrule
\end{tabular}}
\label{tab:quality}
\end{table}

\begin{figure}[t!]
\centerline{
      \includegraphics[width=1\linewidth]{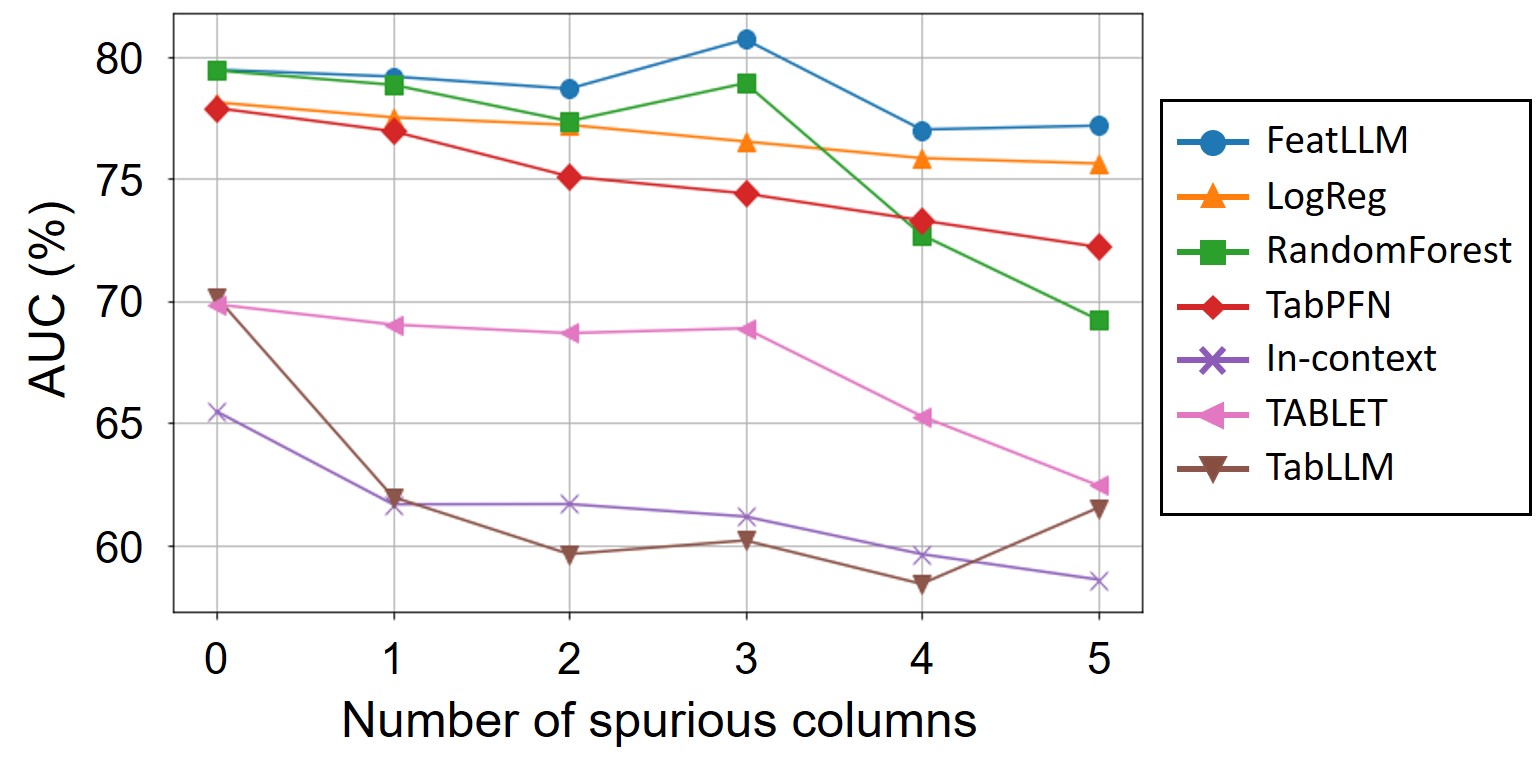}}
      \caption{Visualization of performance impact from spurious correlations. The results exhibit the models' performance (AUC) each time a noisy column from the Adult dataset is added to the original Heart dataset. XGBoost is excluded here due to its lower performance. \looseness=-1
      } 
\label{fig:spurious}
\end{figure}

\cutparagraphup
\paragraph{Effect of adding spurious correlations.}
To ascertain how effectively various methods filter out noisy spurious correlations in a low-shot regime, we conducted an analysis. The experiment utilized the Heart dataset, to which we randomly attach columns from the Adult dataset that are irrelevant to the task of the Heart dataset. We then measure the performance changes of the models while progressively increasing the number of unrelated columns. For a fair comparison, we assume the same 8 shots of labeled samples without considering an unlabeled dataset, thereby excluding methods like SCARF or STUNT from the comparison. \looseness=-1

Figure~\ref{fig:spurious} visualizes the performance changes due to spurious correlations. Compared to baselines, \model{} showed the least performance degradation from spurious correlations. This is likely because our framework forces the consideration of the relationship between the task and features based on prior knowledge when extracting rules, preventing the generation of rules based on irrelevant columns and thereby yielding robust results. Indeed, we observe that rules based on noisy columns accounted for only 1-2\% of the total. Meanwhile, we also observe that \model{} can also learn spurious correlations and misunderstand causal relationships between task and features if columns from closely related data are randomly attached (see Appendix~\ref{sec:spurious-appendix}). \looseness=-1

\begin{figure}[t!]
\includegraphics[width=1\linewidth]{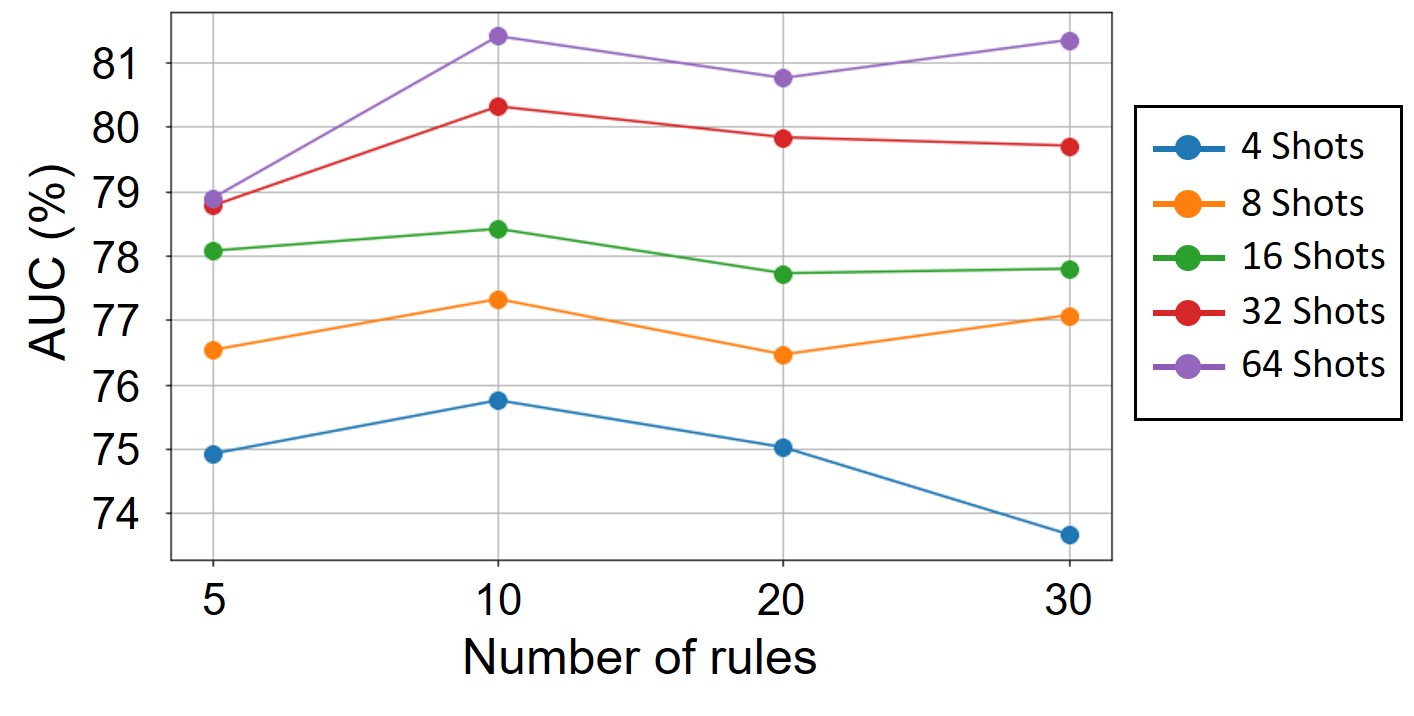}
\caption{Effect of the number of rules per each rule extraction. Averaged AUC over all datasets is reported.}
\label{fig:hyperparameter2}
\end{figure}

\cutparagraphup
\paragraph{Hyper-parameter analysis}
\model{} has two main hyper-parameters. One is the number of ensembles, and the other is how many rules per class to extract during each LLM query. We conducted experiments to investigate the impact of these hyper-parameters on performance. The results showed that performance improves as the number of ensembles increases, but beyond a certain point (i.e., around 20), the performance gains begin to converge (see Fig.~\ref{fig:appendix-hyperparameter1} in Appendix). As for the number of rules, while there wasn't a statistically significant difference, we determined that 10 is the most appropriate number considering the trade-off between prompt length and performance (see Fig.~\ref{fig:hyperparameter2}). We speculate this result is due to more rules enlarging input dimensionality, leading to an increase in information, yet also resulting in overfitting within the low-shot regime.

\section{Conclusion}
We present \model{} which advances the state-of-the-art in LLM-based few-shot tabular learning. \model{} goes beyond using the LLMs for predictions for each sample, and focuses on generating criteria for making predictions as feature engineers. Through rule generation and an ensemble process with bagging, \model{}  requires minimal inference cost, needs only API access without training, and overcomes the constraints to data feature size. \model{}  achieves high prediction performance and is a superior alternative for applying LLM on real-world tabular data in a data-efficient way. \looseness=-1

\model{} is designed to focus on low-shot learing regime only. Our future goal is to expand its capabilities of crafting new features for datasets with a larger number of samples, exploring various feature types beyond rules, and bringing interpretability, which enables its application across a wider array of tasks.  \looseness=-1

\section{Broader Impact}
The proposed framework, \model{}, aims to utilize the prior knowledge of LLM in performing tabular learning tasks to go beyond the achievable performance with standard supervised tabular learning. 
To push the data efficiency in learning, our research makes important contributions to mitigate the issue of overfitting due to limited training samples, by supplementing with prior knowledge. \model{} is particularly beneficial for practitioners in real-world scenarios (such as from finance and healthcare industries) where labeling is challenging due to the high human cost or label collection difficulties, and prior knowledge in LLMs can be potentially very valuable (as the LLMs are pretrained with relevant finance or healthcare related text). 
As one potential important future work direction towards widespread adoption, analyses of the impact of the societal biases and misinformation embedded in the prior knowledge of LLMs would be important, as the prior knowledge would be directly affecting the predictions even dominating over the observed or collected labeled data samples.

\section*{Acknowledgement}
We sincerely thank Scott Yak and Yihe Dong for their valuable feedback and discussion on this work.


\bibliography{main}
\bibliographystyle{icml2024}

\newpage

\appendix
\onecolumn

\section*{Appendix}
\section{Examples of Prompts and Their Corresponding Outputs}
\label{sec:examples}
\subsection{Task Description for Each Dataset}
\label{sec:task_desc}
This section demonstrates creation of prompts for rule extraction for each task, along with the task descriptions we use (see Section~\ref{Sec:prompt_design} - Basic Information Description). The descriptions of the tasks are developed by referring to the descriptions in the original data links and previous works~\cite{hegselmann2023tabllm}, as outlined in the table below.

\begin{table}[h!]
\caption{Description of the tasks associated with each dataset utilized in the prompt. The text in \textcolor{blue}{blue} describes the label information, providing the list of possible answer classes.}
\begin{tabular}{@{}l|l@{}}
\toprule
Data & Task description \\ \midrule
Adult & Does this person earn more than 50000 dollars per year? \textcolor{blue}{Yes or no?} \\
Bank & Does this client subscribe to a term deposit? \textcolor{blue}{Yes or no?} \\
Blood & Did the person donate blood? \textcolor{blue}{Yes or no?} \\
Car & How would you rate the decision to buy this car? \textcolor{blue}{Unacceptable, acceptable, good or very good?} \\
Communities & How high will the rate of violent crimes per 100K population be in this area. \textcolor{blue}{Low, medium, or high?} \\
Credit-g & Does this person receive a credit? \textcolor{blue}{Yes or no?} \\
Diabetes & Does this patient have diabetes? \textcolor{blue}{Yes or no?} \\
Heart & Does the coronary angiography of this patient show a heart disease? \textcolor{blue}{Yes or no?} \\
Myocardial & Does the myocardial infarction complications data of this patient show chronic heart failure? \textcolor{blue}{Yes or no?} \\ 
Cultivars & How high will the grain yield of this soybean cultivar. \textcolor{blue}{Low or high?} \\
NHANES & Predict this person's age group from the given record. \textcolor{blue}{Senior or non-senior?} \\
Sequence-type & What is the type of following sequence? \textcolor{blue}{Arithmetic, geometric, fibonacci, or collatz?} \\
Solution-mix & Given the volumes and concentrations of four solutions, does the percent concentration of the mixed \\
& solution over 0.5? \textcolor{blue}{Yes or no?} \\
\bottomrule
\end{tabular}
\label{Tab:appendix-description}
\end{table}

\newpage

\subsection{Example Prompt for Extracting Rules}
\label{sec:example_rule}
We introduce an example prompt designed for rule extraction in the `Heart' dataset. This prompt utilizes the template illustrated in Figure~\ref{fig:prompt_for_rule}, with relevant information filled in. We assume the 4-shot setting.

\begin{figure}[ht!]
\vspace{-3mm}
\begin{tcolorbox}[enhanced,attach boxed title to top center={yshift=-3mm,yshifttext=-1mm},
  colback=black!0!white, colframe=black!20!white, colbacktitle=black!10!white, coltitle=blue!20!black ]
\footnotesize{
You are an expert. Given the task description and the list of features and data examples, you are extracting conditions for each answer class to solve the task. \\

Task: Does the coronary angiography of this patient show a heart disease? Yes or no? \\

Features: \\
- Age: age of the patient (numerical variable)\\
- Sex: sex of the patient (categorical variable with categories [M, F])\\
- ChestPainType: chest pain type (categorical variable with categories [ATA, NAP, ASY, TA])\\
- RestingBP: resting blood pressure [mm Hg] (numerical variable)\\
- Cholesterol: serum cholesterol [mm/dl] (numerical variable)\\
- FastingBS: fasting blood sugar [1: if FastingBS $>$ 120 mg/dl, 0: otherwise] (numerical variable)\\
- RestingECG: resting electrocardiogram results (categorical variable with categories [Normal, ST, LVH])\\
- MaxHR: maximum heart rate achieved (numerical variable)\\
- ExerciseAngina: exercise-induced angina (categorical variable with categories [N, Y])\\
- Oldpeak: oldpeak = ST [Numeric value measured in depression] (numerical variable)\\
- ST\_Slope: the slope of the peak exercise ST segment (categorical variable with categories [Up, Flat, Down])\\

Examples:  \\
Age is 63. Sex is M. ChestPainType is NAP. RestingBP is 130. Cholesterol is 0. FastingBS is 1. RestingECG is ST. MaxHR is 160. ExerciseAngina is N. Oldpeak is 3.0. ST\_Slope is Flat. \\
Answer: no \\
Age is 39. Sex is M. ChestPainType is ATA. RestingBP is 120. Cholesterol is 204. FastingBS is 0. RestingECG is Normal. MaxHR is 145. ExerciseAngina is N. Oldpeak is 0.0. ST\_Slope is Up. \\
Answer: no \\
Age is 58. Sex is M. ChestPainType is NAP. RestingBP is 160. Cholesterol is 211. FastingBS is 1. RestingECG is ST. MaxHR is 92. ExerciseAngina is N. Oldpeak is 0.0. ST\_Slope is Flat. \\ 
Answer: yes\\ 
Age is 55. Sex is M. ChestPainType is ASY. RestingBP is 160. Cholesterol is 289. FastingBS is 0. RestingECG is LVH. MaxHR is 145. ExerciseAngina is Y. Oldpeak is 0.8. ST\_Slope is Flat. \\
Answer: yes \\

Let's first understand the problem and solve the problem step by step. \\
Step 1. Analyze the causal relationship or tendency between each feature and task description based on general knowledge and common sense within a short sentence. \\
Step 2. Based on the above examples and Step 1's results, infer 10 different conditions per answer, following the format below. The condition should make sense, well match examples, and must match the format for [condition] according to value type.\\

Format for Response: \\
10 different conditions for class ``no": \\
- [Condition] \\
... \\ 

10 different conditions for class ``yes": \\
- [Condition] \\
... \\

Format for [Condition]: \\
For the categorical variable only, \\
- [Feature] is in [List of categories]\\
For the numerical variable only,\\
- [Feature] ($>$ or $>=$ or $<$ or $<=$) [Value]\\
- [Feature] is within range of [Value\_start, Value\_end]\\

Answer: Step 1. 
}
\end{tcolorbox}
\captionof{figure}{Example prompt for rule extraction in the `Heart' dataset}
\label{fig:appendix-prompt_for_rule}            
\end{figure}

\newpage

\subsection{Example Outcome from Rule Extraction}
\label{sec:outcome-rule}
We demonstrate how rules are derived through prompts by showing an example of an outcome. If we input the prompt from the previous subsection (see Fig.~\ref{fig:appendix-prompt_for_rule}), the following result is obtained.

\begin{figure}[ht!]
\begin{tcolorbox}[enhanced,attach boxed title to top center={yshift=-3mm,yshifttext=-1mm},
  colback=black!0!white, colframe=black!20!white, colbacktitle=black!10!white, coltitle=blue!20!black ]
\footnotesize{
Step 1. The relationship between each feature and the task description: \\

- Age: There is a tendency for older age to be associated with a higher risk of heart disease. \\
- Sex: There may be a difference in the prevalence of heart disease between males and females, with males potentially having a higher risk. \\
- ChestPainType: Certain types of chest pain, such as NAP and ASY, may be indicative of heart disease. \\
- RestingBP: Higher resting blood pressure levels may be associated with a higher risk of heart disease. \\
- Cholesterol: Higher levels of serum cholesterol may be associated with a higher risk of heart disease. \\
- FastingBS: Fasting blood sugar levels above 120 mg/dl may be indicative of heart disease. \\
- RestingECG: Abnormal resting electrocardiogram results, such as ST and LVH, may be indicative of heart disease. \\
- MaxHR: Lower maximum heart rate achieved may be associated with a higher risk of heart disease. \\
- ExerciseAngina: The presence of exercise-induced angina may be indicative of heart disease. \\
- Oldpeak: Higher values of oldpeak (ST depression) may be associated with a higher risk of heart disease. \\
- ST\_Slope: The slope of the peak exercise ST segment may provide information about the presence of heart disease.  \\~\\

Step 2. Inferred conditions for each answer class: \\

10 different conditions for class ``no": \\
- Age is within range of [40, 60] \\
- Sex is in [F] \\
- ChestPainType is in [ATA, NAP] \\
- RestingBP ($<$ 140) \\
- Cholesterol ($<$ 200) \\
- FastingBS is in [0] \\
- RestingECG is in [Normal] \\
- MaxHR ($>$ 140) \\
- ExerciseAngina is in [N] \\
- Oldpeak ($<$ 1.0) \\

10 different conditions for class ``yes": \\
- Age is within range of [50, 70] \\
- Sex is in [M] \\
- ChestPainType is in [ASY, TA] \\
- RestingBP ($>$ 140) \\
- Cholesterol ($>$ 200) \\
- FastingBS is in [1] \\
- RestingECG is in [ST, LVH] \\
- MaxHR ($<$ 150) \\
- ExerciseAngina is in [Y] \\
- Oldpeak ($>$ 1.0) \\
}
\end{tcolorbox}
\captionof{figure}{Example outcome from rule extraction in the `Heart' dataset}
\label{fig:appendix-rule_outcome}            
\end{figure}

\newpage

\subsection{Example Prompt for Parsing Rules}
\label{sec:example_parsing}
Utilizing the given rules, we convert them into code capable of transforming each data point into a binary feature. This binary feature indicates whether the sample adheres to the rules associated with its class. Below is an example prompt for parsing rules, employing the same example as in the previous subsection.

\begin{figure}[ht!]
\begin{tcolorbox}[enhanced,attach boxed title to top center={yshift=-3mm,yshifttext=-1mm},
  colback=black!0!white, colframe=black!20!white, colbacktitle=black!10!white, coltitle=blue!20!black ]
\footnotesize{
Provide me a python code for function, given description below. \\

Function name: extracting\_features\_no \\

Input: Dataframe df\_input \\

Input Features: \\
- Age: age of the patient (numerical variable)\\
- Sex: sex of the patient (categorical variable with categories [M, F])\\
- ChestPainType: chest pain type (categorical variable with categories [ATA, NAP, ASY, TA])\\
- RestingBP: resting blood pressure [mm Hg] (numerical variable)\\
- Cholesterol: serum cholesterol [mm/dl] (numerical variable)\\
- FastingBS: fasting blood sugar [1: if FastingBS $>$ 120 mg/dl, 0: otherwise] (numerical variable)\\
- RestingECG: resting electrocardiogram results (categorical variable with categories [Normal, ST, LVH])\\
- MaxHR: maximum heart rate achieved (numerical variable)\\
- ExerciseAngina: exercise-induced angina (categorical variable with categories [N, Y])\\
- Oldpeak: oldpeak = ST [Numeric value measured in depression] (numerical variable)\\
- ST\_Slope: the slope of the peak exercise ST segment (categorical variable with categories [Up, Flat, Down])\\

Output: Dataframe df\_output. Create a new dataframe df\_output. Each column in df\_output refers whether the selected column in df\_input follows the condition (1) or not (0). Be sure that the function code well matches with its feature type (i.e., numerical, categorical). \\

Conditions: \\
- Age is within range of [40, 60] \\
- Sex is in [F] \\
- ChestPainType is in [ATA, NAP] \\
- RestingBP ($<$ 140) \\
- Cholesterol ($<$ 200) \\
- FastingBS is in [0] \\
- RestingECG is in [Normal] \\
- MaxHR ($>$ 140) \\
- ExerciseAngina is in [N] \\
- Oldpeak ($<$ 1.0) \\

Wrap only the function part with $<$start$>$ and $<$end$>$, and do not add any comments, descriptions, and package importing lines in the code.
}
\end{tcolorbox}
\captionof{figure}{Example prompt for parsing rules in the `Heart' dataset}
\label{fig:appendix-parse_rules}            
\end{figure}

\newpage

\subsection{Example Outcome After Parsing}
\label{sec:example_parse_outcome}
The result of the parsing request is exemplified as follows.

\begin{figure}[ht!]
\begin{tcolorbox}[enhanced,attach boxed title to top center={yshift=-3mm,yshifttext=-1mm},
  colback=black!0!white, colframe=black!20!white, colbacktitle=black!10!white, coltitle=blue!20!black ]
\footnotesize{
$<$start$>$\\
def extracting\_features\_no(df\_input): \\

\hangindent=0.7cm
\parindent=0.7cm df\_output = pd.DataFrame() \\
df\_output[`Age'] = df\_input[`Age'].apply(lambda x: 1 if x $>=$ 40 and x $<=$ 60 else 0)\\
df\_output[`Sex'] = df\_input[`Sex'].apply(lambda x: 1 if x == `F' else 0)\\
df\_output[`ChestPainType'] = df\_input[`ChestPainType'].apply(lambda x: 1 if x in [`ATA', `NAP'] else 0)\\
df\_output[`RestingBP'] = df\_input[`RestingBP'].apply(lambda x: 1 if x $<$ 140 else 0)\\
df\_output[`Cholesterol'] = df\_input[`Cholesterol'].apply(lambda x: 1 if x $<$ 200 else 0)\\
df\_output[`FastingBS'] = df\_input[`FastingBS'].apply(lambda x: 1 if x == 0 else 0)\\
df\_output[`RestingECG'] = df\_input[`RestingECG'].apply(lambda x: 1 if x == `Normal' else 0)\\
df\_output[`MaxHR'] = df\_input[`MaxHR'].apply(lambda x: 1 if x $>$ 140 else 0)\\
df\_output[`ExerciseAngina'] = df\_input[`ExerciseAngina'].apply(lambda x: 1 if x = `N' else 0)\\
df\_output[`Oldpeak'] = df\_input[`Oldpeak'].apply(lambda x: 1 if x $<$ 1.0 else 0)\\
    
return df\_output \\
\hangindent=0cm
$<$end$>$
}
\end{tcolorbox}
\captionof{figure}{Example outcome after parsing rules in the `Heart' dataset}
\label{fig:appendix-parse_outcomes}            
\end{figure}

\section{Hyper-parameter analysis}
\label{sec:hyper-parameter}
\begin{figure}[h!]
\centering
\includegraphics[width=0.48\textwidth]{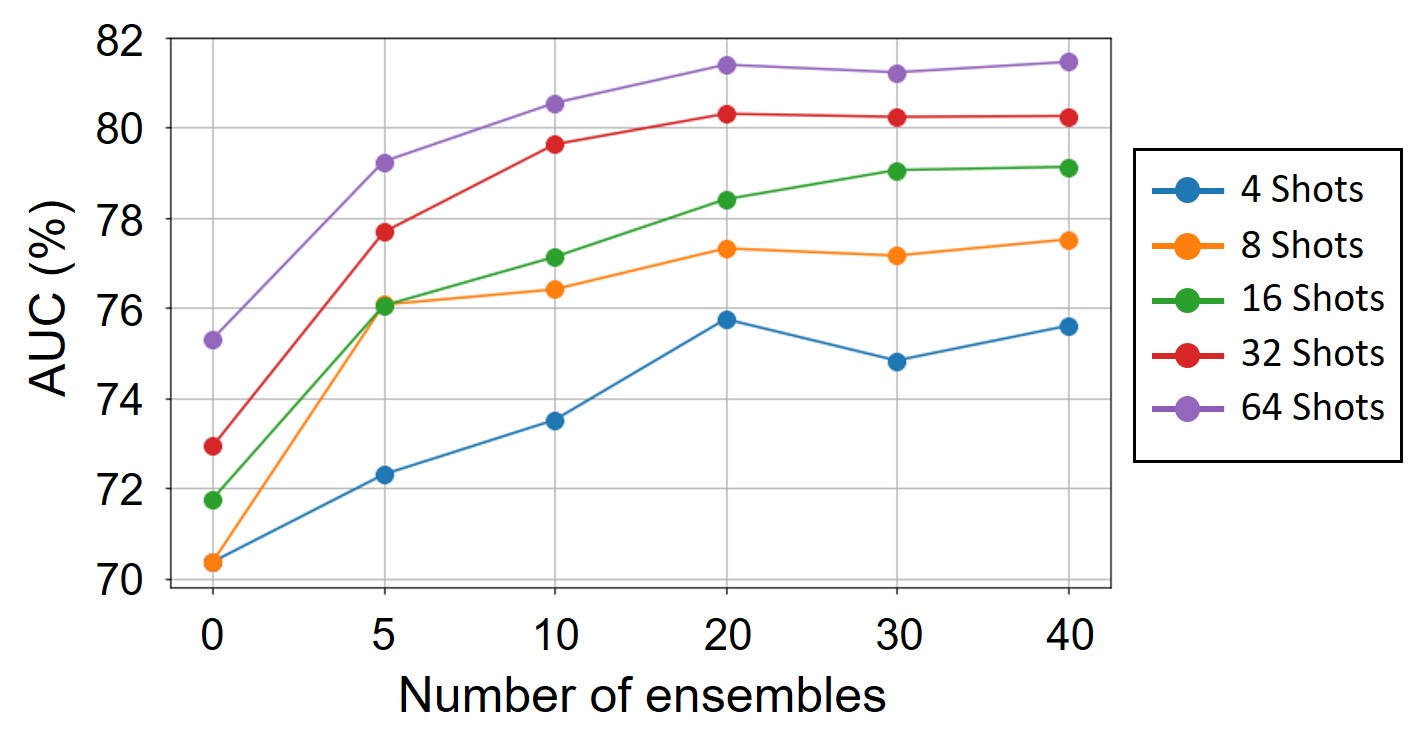}
\captionof{figure}{Effect of the number of ensembles. Averaged AUC over all datasets is reported.}
\label{fig:appendix-hyperparameter1}
\end{figure}

\model{} has two main hyper-parameters. One is the number of ensembles, and the other is how many rules per class to extract during each LLM query. We conducted experiments to investigate the impact of these hyper-parameters on performance. The results showed that performance improves as the number of ensembles increases, but beyond a certain point (i.e., around 20), the performance gains begin to converge (see Fig.~\ref{fig:appendix-hyperparameter1}). For an analysis of the impact of the number of rules, please refer to Section~\ref{sec:analysis}. \looseness=-1

\newpage

\section{Baseline Implementation Details}
\label{sec:baselines}
In this section, we provide a detailed information on the implementation of baselines. 
In our study, we adhere to the settings and parameters outlined in the original papers for various baselines. During model training, the dataset was divided, allocating 20\% as the test set. This division was carried out using stratified sampling to ensure a representative distribution of classes. From the remaining data, $k$-shot samples were selected, constructing a balanced training set. We repeated experiments multiple times with different randomness on the choice of training/test sets. 

For conventional machine learning methods, including logistic regression (LogReg), RandomForest, and XGBoost, we determined their parameters using Grid-search combined with $k$-shot cross validation. Following the previous work~\cite{hegselmann2023tabllm}, the search space of each parameter is described in Tables below.

\begin{table}[h!]
\centering
\parbox{.45\linewidth}{
\centering
\begin{tabular}{@{}l|l@{}}
\toprule
Parameter & Search space \\ \midrule
penalty & l1, l2 \\
C & 100, 10, 1, 1e-1, 1e-2, 1e-3, 1e-4, 1e-5 \\ \bottomrule
\end{tabular}
\label{tab:appendix-logreg}
\caption{Hyperparameter search space for LogReg}}
\parbox{.50\linewidth}{
\centering
\begin{tabular}{@{}l|l@{}}
\toprule
Parameter & Search space \\ \midrule
bootstrap & True, False \\
max\_depth & 2, 4, 6, 8, 10, 12 \\
n\_estimators & 2, 4, 8, 16, 32, 64, 128, 256 \\ \bottomrule
\end{tabular}
\caption{Hyperparameter search space for RandomForest}}
\end{table}

\begin{table}[h!]
\centering
\begin{tabular}{@{}l|l@{}}
\toprule
Parameter & Search space \\ \midrule
max\_depth & 2, 4, 6, 8, 10, 12 \\
alpha & 1e-8, 1e-7, 1e-6, 1e-5, 1e-4, 1e-3, 1e-2, 1e-1, 1 \\
lambda & 1e-8, 1e-7, 1e-6, 1e-5, 1e-4, 1e-3, 1e-2, 1e-1, 1 \\
eta & 0.01, 0.03, 0.1, 0.3 \\ \bottomrule
\end{tabular}
\caption{Hyperparameter search space for XGBoost}
\end{table}

For baseline models that incorporate unlabeled data in training, such as SCARF and STUNT, any data not used in the test set or as $k$-shot training samples is treated as the unlabeled dataset. To ensure that there are no missing values in the training and test data, columns representing more than 20\% of the entire dataset were removed, similarly following the literature~\cite{kang2013prevention,jakobsen2017and}. Subsequently, any rows containing missing values were also eliminated. 
For TabPFN, we utilized the official GitHub repository\footnote{https://github.com/automl/TabPFN} to reproduce results, maintaining the default parameter settings as specified.

Regarding LLM-based baselines, we utilized GPT-3.5 for in-context learning, while T0 was chosen for the TabLLM, in line with their original usage settings. To calculate the AUC for the in-context learning and TABLET with the GPT-3.5 API, we performed the inference process three times, introducing variability with a nonzero temperature setting (i.e., 1). The results were then averaged to estimate the probabilities for each class. In contrast, for the TabLLM, we employed T0, utilizing its publicly available checkpoint. We directly calculated the probability of the class name token to estimate the class probabilities.

\newpage

\section{Diversity of Features Generated by \model{}}
\label{sec:diversity}
In our model, to guarantee diversity of features (i.e., rules) generated from the LLM in each iteration, we set the temperature of LLMs to a value greater than 0 (i.e., 0.5) to introduce randomness into the generation, applied bagging strategies (i.e., feature and sample bagging), and shuffled the order of training samples in the prompt for each iteration. To empirically verify the diversity of features, we conducted the following analysis: \vspace{-3mm}
\begin{enumerate}[noitemsep]
    \item Given two sets of rules from different iterations, we applied each set's rules across the dataset to generate binary features. \looseness=-1
    \item We used the Hungarian Maximum Matching Algorithm~\cite{kuhn1955hungarian} to create a bipartite graph connecting pairs of binary features between the two sets with the highest correlation.
    \item The average correlation of the selected pairs was calculated.
    \item This process (steps 1-3) was repeated for all pairs of iterations to determine the average and variance of the correlations. \vspace{-3mm}
\end{enumerate}
We conducted an ablation study to assess the effects of three components (i.e., temperature, bagging, instance shuffling) on enhancing rule diversity. Table~\ref{tab:appendix-diversity} below presents the results of average correlation across iterations, where a lower correlation signifies a more diverse set of extracted rules. We observed that the average correlation was 0.368 with a variance of 0.341, indicating a generally low correlation and confirming the generation of diverse rules. Additionally, we noted an increase in correlation when each of these components was removed, demonstrating that each component contributes to diversity. \looseness=-1

\begin{table}[!ht]
\centering
\caption{Ablation study summaries on rule diversity.}
\scalebox{1}{
\begin{tabular}{l|c}
\toprule
Ablations & Correlation between rule sets across iterations \\ \midrule
All strategies & 0.368 $\pm$ 0.341 \\
without bagging & 0.384 $\pm$ 0.327 \\
without instance shuffling & 0.531 $\pm$ 0.290 \\
with zero temperature & 0.414 $\pm$ 0.373 \\ 
\bottomrule
\end{tabular}}
\label{tab:appendix-diversity}
\end{table}

\section{Prior Knowledge vs. Training Samples.}
\model{} simultaneously utilizes prior knowledge from LLMs and information extracted from training samples to derive rules. This raises the question: which source of knowledge does the framework prioritize when generating rules? To investigate this, we conducted a simple experiment. Keeping the training data unchanged, we reversed its labels and evaluated the resulting rule set. The experiment used Adult, and to focus solely on the rule set, we removed ensemble and weight tuning processes from the evaluation. Table~\ref{tab:appendix-flipped} shows the results for rule sets evaluated with both the original and flipped data. Despite the label reversal, our framework tends to more focused on prior knowledge. However, we also observed a significant increase in the standard deviation of AUC, indicating variability in each trial, with some prioritizing training samples and others prior knowledge, seemingly at random. These findings underscore the importance of using ensemble process to cover both scenarios effectively.

\begin{table}[h!]
\centering
\caption{Comparison of AUC between original data and flipped data whose labels in training samples are reversely flipped. Adult dataset is utilized.}
\scalebox{1}{
\begin{tabular}{@{}c|ccccc@{}}
\toprule
Shot & 4 & 8 & 16 & 32 & 64 \\ \midrule
Original & 83.66$\pm$6.76 & 83.48$\pm$7.21 & 84.51$\pm$2.64 & 84.16$\pm$5.32 & 84.07$\pm$3.38 \\
Flipped & 63.91$\pm$22.54 & 71.35$\pm$18.53 & 66.81$\pm$18.57 & 65.94$\pm$18.57 & 56.14$\pm$17.39 \\ \bottomrule
\end{tabular}}
\label{tab:appendix-flipped}
\end{table}

\newpage

\section{Error Rates of Automated Parsing via LLM}
\model{} utilizes rules extracted through an LLM for parsing and converting data. However, since the conversion code is derived from the LLM, there are some cases where the generated code does not function properly and fails. We have conducted experiments and recorded how frequently such failures occur across different datasets. According to Table~\ref{tab:appendix-errors}, the error rate varies per dataset, with higher failure rates often seen in data containing many categorical features. This is because GPT-3.5 sometimes makes syntactic errors in handling categorical columns in the pandas library, and the probability of errors increases with more rules pertaining to categorical variables. We believe that using a model specialized in code generation or a higher-performance model (e.g., GPT-4) could significantly reduce this error rate.

\begin{table}[h!]
\centering
\caption{Error rate (\%) of automated parsing for each dataset.}
\begin{tabular}{@{}lc@{}}
\toprule
Data & Error rate \\ \midrule
Adult & 14.67 \\
Bank & 35.33 \\
Blood & 1.33 \\
Car & 1.67 \\
Communities & 1.00 \\
Credit-g & 40.33 \\
Diabetes & 1.67 \\
Heart & 2.67 \\
Myocardial & 13.00 \\
Cultivars & 12.85 \\
NHANES & 0.00 \\
Sequence-type & 35.36 \\
Solution-mix & 18.43 \\ \bottomrule
\end{tabular}
\label{tab:appendix-errors}
\end{table}

\section{Handling missing data.}
Given missing values in data, \model{} requires an additional processing step to handle them. Specifically, when creating binary features through rules, we need to decide how to handle these missing values. We have compared three strategies to address this:
\begin{itemize}
    \item \textsf{Fill zero}: Assuming that the rules related to the missing value are not satisfied, fill these with zero.
    \item \textsf{Fill 0.5}: Since the absence of a value leaves it unclear whether the rule is satisfied or not, fill the value as 0.5.
    \item \textsf{Imputation}: Using the binary matrix of the training data, impute the missing values via MICE~\cite{van2011mice} considering the relationships between the rules.
\end{itemize}

Table~\ref{tab:appendix-missing} presents a summary of performance comparisons based on different strategies in relation to the missing ratio. Compared to other strategies, imputation demonstrated the lowest performance degradation. We speculate that the effectiveness of imputation, even with a limited number of shots, is due to the reduced complexity of imputation, a result of converting data to binary form through rules.

\begin{table}[h!]
\centering
\caption{Missing value analysis. Averaged changes of AUC with standard error over 13 datasets across different missing ratios are reported. \looseness=-1}
\setlength{\tabcolsep}{3.5pt}

\scalebox{1.05}{
\begin{tabular}{@{}l|cllll@{}}
\toprule
Missing ratio & 0 & \multicolumn{1}{c}{0.05} & \multicolumn{1}{c}{0.1} & \multicolumn{1}{c}{0.15} & \multicolumn{1}{c}{0.2} \\ \midrule
Fill zero & \multirow{3}{*}{78.64} & -0.49$\pm$0.23 & -1.57$\pm$0.21 & -1.75$\pm$0.28 & -2.15$\pm$0.44 \\
Fill 0.5 &  & -0.53$\pm$0.27 & -1.55$\pm$0.22 & -1.72$\pm$0.34 & -2.07$\pm$0.50 \\
Imputation &  & -0.29$\pm$0.20 & -1.21$\pm$0.27 & -1.41$\pm$0.29 & -1.88$\pm$0.28 \\ \bottomrule
\end{tabular}}
\label{tab:appendix-missing}
\end{table}

\newpage

\section{Varying LLM backbones}
\label{sec:backbones}
To investigate the impact of using different LLMs on our framework, given their distinct prior knowledge and reasoning abilities from being trained on various text corpora, we measured performance using not only GPT-3.5 but also the PaLM 2 Text-Unicorn model as a backbone for comparison. The results in Table~\ref{tab:appendix-differentLLM} indicate that certain tasks show greater improvement with specific LLMs.

\begin{table}[h!]
\caption{Effect of varying LLM backbones: GPT-3.5 vs. PaLM 2-Unicorn}
\centering
\scalebox{0.9}{
\begin{tabular}{@{}l|c|cc@{}}
\toprule
Data & Shot & \model{} (GPT-3.5) & \model{} (PaLM 2-Unicorn) \\ \midrule
Adult & 4 & \textbf{86.68$\pm$0.86} & 82.99$\pm$2.86 \\
 & 8 & \textbf{87.89$\pm$0.06} & 83.24$\pm$0.87 \\
 & 16 & \textbf{87.54$\pm$0.50} & 83.19$\pm$1.89 \\
 & 32 & \textbf{87.09$\pm$0.58} & 83.93$\pm$3.50 \\
 & 64 & \textbf{87.77$\pm$0.31} & 86.18$\pm$1.79 \\ \midrule
Bank & 4 & \textbf{70.45$\pm$3.69} & 60.89$\pm$6.07 \\
 & 8 & \textbf{75.85$\pm$6.66} & 65.44$\pm$8.26 \\
 & 16 & \textbf{78.41$\pm$1.08} & 71.86$\pm$1.33 \\
 & 32 & \textbf{78.37$\pm$4.50} & 76.36$\pm$4.45 \\
 & 64 & \textbf{81.18$\pm$6.17} & 79.66$\pm$4.22 \\ \midrule
Blood & 4 & \textbf{68.34$\pm$7.48} & 60.30$\pm$13.99 \\
 & 8 & \textbf{70.37$\pm$3.23} & 67.67$\pm$1.25 \\
 & 16 & \textbf{70.07$\pm$5.19} & 67.54$\pm$3.39 \\
 & 32 & \textbf{71.13$\pm$4.38} & 69.61$\pm$2.07 \\
 & 64 & \textbf{71.04$\pm$4.36} & 69.83$\pm$5.03 \\ \midrule
Car & 4 & \textbf{72.69$\pm$1.52} & 66.92$\pm$7.86 \\
 & 8 & \textbf{73.26$\pm$1.46} & 71.59$\pm$1.53 \\
 & 16 & \textbf{79.43$\pm$1.24} & 74.40$\pm$1.87 \\
 & 32 & \textbf{85.01$\pm$1.36} & 78.10$\pm$1.67 \\
 & 64 & \textbf{86.78$\pm$0.90} & 80.22$\pm$3.25 \\ \midrule
Communities & 4 & \textbf{75.39$\pm$5.05} & 73.87$\pm$5.56 \\
 & 8 & 76.59$\pm$1.25 & \textbf{78.57$\pm$1.46} \\
 & 16 & 76.25$\pm$0.64 & \textbf{77.87$\pm$2.09} \\
 & 32 & 76.85$\pm$1.38 & \textbf{78.92$\pm$2.17} \\
 & 64 & 78.62$\pm$1.39 & \textbf{79.38$\pm$1.66} \\ \midrule
Credit-g & 4 & \textbf{55.94$\pm$1.10} & 54.00$\pm$5.91 \\
 & 8 & \textbf{57.42$\pm$3.10} & 54.20$\pm$3.50 \\
 & 16 & 56.60$\pm$2.22 & \textbf{57.35$\pm$6.20} \\
 & 32 & \textbf{61.79$\pm$10.25} & 58.49$\pm$5.18 \\
 & 64 & \textbf{66.43$\pm$2.90} & 66.09$\pm$7.29 \\ \midrule
Diabetes & 4 & \textbf{80.28$\pm$0.75} & 78.40$\pm$2.20 \\
 & 8 & \textbf{79.38$\pm$1.66} & 78.27$\pm$1.48 \\
 & 16 & \textbf{80.15$\pm$1.35} & 79.99$\pm$1.45 \\
 & 32 & \textbf{80.06$\pm$1.18} & 79.92$\pm$1.91 \\
 & 64 & \textbf{80.91$\pm$1.62} & 79.20$\pm$0.43 \\ \midrule
Heart & 4 & \textbf{75.66$\pm$4.59} & 75.12$\pm$3.97 \\
 & 8 & 79.46$\pm$2.16 & \textbf{82.07$\pm$10.08} \\
 & 16 & 83.71$\pm$1.88 & \textbf{86.18$\pm$3.83} \\
 & 32 & 87.19$\pm$3.66 & \textbf{87.99$\pm$3.26} \\
 & 64 & 88.08$\pm$4.11 & \textbf{88.28$\pm$3.34} \\ \midrule
Myocardial & 4 & 52.87$\pm$3.44 & \textbf{59.48$\pm$7.25} \\
 & 8 & 56.22$\pm$1.64 & \textbf{61.94$\pm$2.35} \\
 & 16 & 55.32$\pm$9.15 & \textbf{62.17$\pm$3.58} \\
 & 32 & 60.02$\pm$4.02 & \textbf{60.72$\pm$5.18} \\
 & 64 & \textbf{61.47$\pm$3.91} & 61.25$\pm$5.60 \\ \bottomrule
\end{tabular}}
\label{tab:appendix-differentLLM}
\end{table}

\newpage

\section{Additional Experiments on Effect of Adding Spurious Correlations}
\label{sec:spurious-appendix}

\begin{figure}[h!]
\centerline{
      \includegraphics[width=0.65\linewidth]{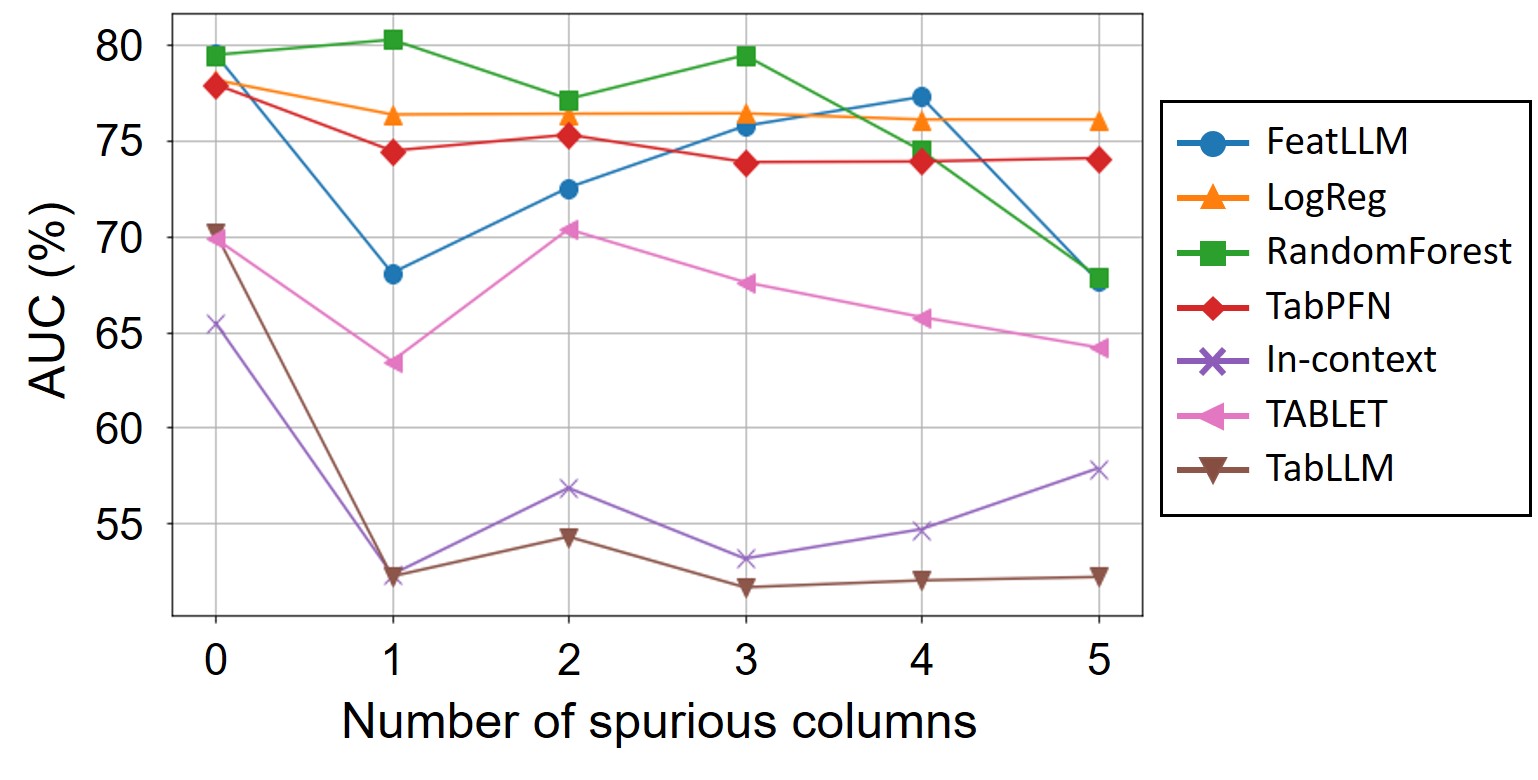}}
      \caption{Visualization of performance impact from spurious correlations. The results exhibit the models' performance (AUC) each time a noisy column from the Myocardial dataset is added to the original Heart dataset. XGBoost is excluded here due to its lower performance. \looseness=-1
      } 
\label{fig:appendix-spurious}
\end{figure}

In the previous section~\ref{sec:analysis}, ``Effect of adding spurious correlations," we experimented with randomly mixing irrelevant columns from the Adult dataset into the Heart dataset to test for the removal of spurious correlations. In this section, we explored whether similar effects occur when columns from the Myocardial dataset, which is in a similar domain to the Heart dataset, are added. According to Figure~\ref{fig:appendix-spurious}, unlike results in our main experiment, \model{} showed unstable performance. This instability arises because many columns from the Myocardial dataset are relevant to the actual task of the Heart dataset. When these columns are combined with random values, \model{} tends to extract more rules based on these columns, leading to overfitting on noisy information. Indeed, we observed that rules based on noisy columns represented 7-8\% of the total, significantly more than the 1-2\% in the original experiments with the Adult dataset. It can also be observed that other LLM-based methodologies experience a significant decrease in performance. However, as the number of training shots increases (e.g., 64), the linear model will automatically measure the importance, which can also reduce the weight given to irrelevant columns, thereby decreasing such instability.

\newpage

\section{Qualitative Analysis}
\label{sec:qualitative}
Finally, to verify whether the rules deemed important by our framework are indeed significant from the perspective of the actual task definition, we conducted a qualitative analysis. We extracted the rules with the highest weights based on a linear model, selecting three for each data and answer class, and reported them in Table~\ref{tab:appendix_Qualitative}. It was observed that each rule intuitively aligns well with the task description.

{\small
\begin{longtable}{lcl}
\caption{The top-3 rules for each data and answer class, ranked by their measured importance. Importance was determined based on the weights of a trained linear model, with rules extracted in descending order of importance.}
\label{tab:appendix_Qualitative} \\
\toprule
Data & Label & Top-3 rules \\ \midrule
\endfirsthead
\endhead
\hline
\endfoot
\endlastfoot
Adult & No & educational-num \textless{}= 9 \\
 &  & capital-gain \textless{}= 0 \\
 &  & marital-status in {[}Never-married, Widowed, Divorced, Separated, Married-spouse-absent{]} \\
 & Yes & educational-num \textgreater 10 \\
 &  & capital-gain \textgreater 0 \\
 &  & occupation in {[}Machine-op-inspct, Protective-serv, Adm-clerical, Armed-Forces{]} \\ \midrule
Bank & No & contact = unknown \\
 &  & age is within {[}18, 30{]} \\
 &  & education in {[}secondary, unknown{]} \\
 & Yes & balance \textgreater 1000 \\
 &  & loan = no \\
 &  & age is within {[}40, 60{]} \\ \midrule
Blood & No & frequency \textless{}=2 \\
 &  & monetary \textless{}= 500 \\
 &  & time \textless{}= 16 \\
 & Yes & recency \textgreater 16 \\
 &  & frequency \textgreater 2 \\
 &  & monetary \textgreater 500 \\ \midrule
Car & Very good & doors in {[}4, 5more{]} \\
 &  & safety = high \\
 &  & lug\_boot = big \\
 & Good & maint = low \\
 &  & safety = high \\
 &  & buying in {[}medium, low{]} \\
 & Acceptable & maint = high \\
 &  & doors = 2 \\
 &  & buying = very high \\
 & Unacceptable & safety = low \\
 &  & buying = very high \\
 &  & doors = 2 \\ \midrule
Communities & High & NumIlleg \textgreater{}= 0.02 \\
 &  & PctHousOwnOcc \textless{}= 0.4 \\
 &  & RentLowQ \textgreater{}= 0.6 \\
 & Medium & MedOwnCostPctIncNoMtg is within {[}0.3, 0.6{]} \\
 &  & PctBSorMore is within {[}0.3, 0.6{]} \\
 &  & PctHousOwnOcc is within {[}0.4, 0.8{]} \\
 & Low & Numllleg \textless{}= 0.01 \\
 &  & PctImmigRec8 \textless{}= 0.2 \\
 &  & PctHousOwnOcc \textgreater{}= 0.8 \\ \midrule 
Credit-g & No & existing\_credits \textgreater 2 \\
 &  & property\_magnitude in {[}life insurance, no known property{]} \\
 &  & housing in {[}rent, for free{]} \\
 & Yes & num\_dependents \textless{}= 1 \\
 &  & housing not in {[}rent, for free{]} \\
 &  & existing\_credits \textless{}= 2 \\ \midrule
Diabetes & No & Glucose \textless 100 \\
 &  & BMI \textless 25 \\
 &  & Glucose \textless 110 \& BloodPressure \textless 70 \\
 & Yes & BMI \textgreater{}= 30 \& Age \textgreater{}= 40 \\
 &  & SkinThickness \textgreater{}= 20 \& Age \textgreater{}= 45 \\
 &  & Insulin \textgreater{}= 150 \\ \midrule
Heart & No & Cholesterol \textless 200 \\
 &  & Age \textless 50 \\
 &  & ExerciseAngina = N \\
 & Yes & Age \textgreater{}= 50 \\
 &  & ExerciseAngina = Y \\
 &  & FastingBS = 1 \\ \midrule
Myocardial & No & ZSN\_A = 'there is no chronic heart failure' \\
 &  & GEPAR\_S\_n = yes \\
 &  & lat\_im = 'QRS has no changes' \\
 & Yes & ZSN\_A = 'I stage' \\
 &  & Iat\_im = 'QRS is like Qr-complex' \\
 &  & n\_r\_ecg\_p\_04 = no \\ \midrule
Cultivars & High & Cultivar in {[}`ELISA IPRO', `MANU IPRO', `TMG 22X83I2X', `83IX84RSF I2X', `MANU IPRO'{]} \\
 &  & NGL \textgreater{}= 2.56 \\
 &  & NLP \textgreater{}= 50 \\
 & Low & Cultivar in {[}`BRASMAX BÔNUS IPRO', `GNS7700 IPRO', `MONSOY M8606I2X'{]} \\
 &  & Cultivar in {[}`97Y97 IPRO', `NEO 760 CE', `MONSOY M8606I2X', `GNS7700 IPRO'{]} \\
 &  & Cultivar in {[}`NEO 760 CE', `FTR 3868 IPRO', `GNS7700 IPRO', `MONSOY M8606I2X'{]} \\ \midrule
NHANES & Non-senior & RIDAGEYR \textless 60 \\
 &  & LBXGLT \textless 150 \\
 &  & BMXBMI \textless 30 \\
 & Senior & RIDAGEYR \textgreater{}= 60 \\
 &  & LBXGLT \textgreater 150 \\
 &  & BMXBMI \textgreater = 30 \\ \midrule
Sequence-type & Arithmetic & (num2 + (num2 - num1)) == num3 \\
 &  & (num3 + (num3 - num2)) == num4 \\
 &  & (num4 + (num4 - num3)) == num5 \\
 &  & (num3 - (num2 - num1)) == num2 \\
 & Geometric & (num1 * (num3 / num2)) == num2 \\
 &  & (num2 * (num4 / num3)) == num3 \\
 &  & (num3 * (num2 / num1)) == num4 \\
 & Fibonacci & (num1 + num2) == num3 \\
 &  & (num2 + num3) == num4 \\
 &  & (num3 + num4) == num5 \\
 & Collatz & num2 \% 2 == 0 \\
 &  & num4 \% 2 == 0 \\
 &  & num1 \textgreater \ num2 \\ \midrule
Solution-mix & No & (Concentration1 * Volume1 + ... + Concentration4 * Volume4) / (Volume1 + ... + Volume4) \textless{}= 0.5 \\
 &  & Concentration1 \textless{}= 0.1 \\
 &  & Concentration3 \textless{}= 0.3 and Concentration4 \textless{}= 0.3 \\
 & Yes & (Concentration1 * Volume1 + ... + Concentration4 * Volume4) / (Volume1 + ... + Volume4) \textgreater 0.5 \\
 &  & Volume1 \textless{}= 0.5 \\
 &  & Concentration1 \textgreater 0.5 \\
 \bottomrule
\end{longtable}}

\section{Full Results}
\label{sec:full}
We here present full results which were not included in the main manuscript due to space constraints.

\subsection{Main Evaluation}
\begin{table}[h!]
\centering
\caption{Evaluation results, showing AUC across 13 datasets. Best performances are bolded, and our framework's performances, when second-best, are underlined. Failed cases are represented as `N/A'. \looseness=-1}
\setlength{\tabcolsep}{2.5pt}
\scalebox{0.735}{
\begin{tabular}{@{}lccccccccccc@{}}
\toprule
Data & Shot & LogReg & XGBoost & RandomForest & SCARF & TabPFN & STUNT & In-context & TABLET & TabLLM & Ours \\ \midrule
Adult & 4 & 72.10$\pm$12.30 & 50.00$\pm$0.00 & 53.65$\pm$24.42 & 58.34$\pm$15.42 & 60.89$\pm$23.28 & 67.43$\pm$29.61 & 77.51$\pm$5.24 & 75.29$\pm$12.24 & 83.57$\pm$2.69 & \textbf{86.68$\pm$0.86} \\
 & 8 & 76.02$\pm$3.37 & 59.19$\pm$6.92 & 74.20$\pm$7.39 & 72.42$\pm$8.95 & 70.42$\pm$9.96 & 82.16$\pm$6.93 & 79.30$\pm$2.89 & 77.56$\pm$7.56 & 83.52$\pm$4.30 & \textbf{87.89$\pm$0.06} \\
 & 16 & 75.20$\pm$5.10 & 60.68$\pm$13.92 & 70.96$\pm$13.46 & 75.63$\pm$9.56 & 70.34$\pm$9.96 & 80.57$\pm$10.93 & 79.50$\pm$4.57 & 79.74$\pm$5.64 & 83.23$\pm$2.45 & \textbf{87.54$\pm$0.50} \\
 & 32 & 82.02$\pm$3.76 & 74.49$\pm$13.78 & 69.90$\pm$13.97 & 75.80$\pm$10.21 & 68.77$\pm$8.09 & 78.08$\pm$15.15 & 81.89$\pm$4.04 & 78.08$\pm$6.70 & 82.60$\pm$4.14 & \textbf{87.09$\pm$0.58} \\
 & 64 & 81.94$\pm$5.16 & 79.56$\pm$2.18 & 82.36$\pm$2.73 & 80.89$\pm$2.82 & 78.16$\pm$4.43 & 86.01$\pm$0.16 & N/A & N/A & 84.88$\pm$0.97 & \textbf{87.77$\pm$0.31} \\ \hline
Bank & 4 & 63.70$\pm$3.87 & 50.00$\pm$0.00 & 60.59$\pm$3.90 & 58.53$\pm$5.49 & 63.19$\pm$11.60 & 56.34$\pm$12.82 & 61.38$\pm$1.30 & 58.11$\pm$6.29 & 62.51$\pm$8.95 & \textbf{70.45$\pm$3.69} \\
 & 8 & 72.52$\pm$3.21 & 58.78$\pm$10.54 & 61.74$\pm$9.91 & 55.28$\pm$11.88 & 62.81$\pm$7.84 & 63.01$\pm$8.78 & 69.57$\pm$13.35 & 69.08$\pm$6.00 & 63.19$\pm$5.79 & \textbf{75.85$\pm$6.66} \\
 & 16 & 77.51$\pm$3.09 & 70.34$\pm$5.86 & 65.67$\pm$10.43 & 65.81$\pm$1.79 & 73.79$\pm$2.21 & 69.85$\pm$0.95 & 69.76$\pm$8.55 & 69.40$\pm$11.28 & 63.73$\pm$6.43 & \textbf{78.41$\pm$1.08} \\
 & 32 & \textbf{79.63$\pm$3.57} & 76.25$\pm$1.26 & 74.29$\pm$5.32 & 68.45$\pm$1.06 & 77.71$\pm$3.56 & 71.64$\pm$1.65 & 66.93$\pm$5.67 & 73.61$\pm$9.28 & 66.51$\pm$3.92 & {\ul 78.37$\pm$4.50} \\
 & 64 & \textbf{82.27$\pm$1.61} & 81.92$\pm$1.00 & 79.55$\pm$3.19 & 68.28$\pm$3.97 & 82.14$\pm$2.28 & 72.26$\pm$1.62 & N/A & N/A & 70.83$\pm$3.43 & 81.18$\pm$6.17 \\ \hline
Blood & 4 & 56.79$\pm$26.02 & 50.00$\pm$0.00 & 48.50$\pm$12.82 & 56.22$\pm$21.00 & 58.72$\pm$19.16 & 48.57$\pm$6.04 & 56.30$\pm$12.43 & 56.45$\pm$15.45 & 55.87$\pm$13.49 & \textbf{68.34$\pm$7.48} \\
 & 8 & 68.51$\pm$5.16 & 59.97$\pm$1.36 & 63.43$\pm$11.03 & 65.77$\pm$5.00 & 66.30$\pm$10.01 & 60.00$\pm$4.84 & 58.99$\pm$10.12 & 56.37$\pm$11.56 & 66.01$\pm$9.25 & \textbf{70.37$\pm$3.23} \\
 & 16 & 68.30$\pm$6.16 & 63.28$\pm$7.62 & 65.98$\pm$6.49 & 66.27$\pm$5.04 & 64.14$\pm$6.80 & 54.76$\pm$4.53 & 56.59$\pm$5.21 & 60.62$\pm$4.13 & 65.14$\pm$7.55 & \textbf{70.07$\pm$5.19} \\
 & 32 & 67.39$\pm$4.46 & 66.41$\pm$6.37 & 63.46$\pm$4.43 & 69.71$\pm$6.24 & 68.65$\pm$4.37 & 59.87$\pm$3.72 & 58.69$\pm$1.53 & 57.94$\pm$4.16 & 69.95$\pm$3.39 & \textbf{71.13$\pm$4.38} \\
 & 64 & 71.76$\pm$2.56 & 69.46$\pm$2.96 & 68.83$\pm$5.61 & 72.75$\pm$4.36 & \textbf{73.88$\pm$1.97} & 61.75$\pm$2.19 & 65.79$\pm$2.05 & 63.47$\pm$7.36 & 70.88$\pm$1.58 & 71.04$\pm$4.36 \\ \hline
Car & 4 & 62.38$\pm$4.13 & 50.00$\pm$0.00 & 58.75$\pm$3.82 & 62.52$\pm$3.80 & 58.14$\pm$4.15 & 61.32$\pm$3.83 & 62.47$\pm$2.47 & 60.21$\pm$4.81 & \textbf{85.82$\pm$3.65} & {\ul 72.69$\pm$1.52} \\
 & 8 & 72.05$\pm$1.20 & 64.00$\pm$3.57 & 65.49$\pm$6.29 & 72.23$\pm$2.59 & 63.95$\pm$4.35 & 67.86$\pm$0.49 & 67.57$\pm$3.44 & 65.53$\pm$8.00 & \textbf{87.43$\pm$2.56} & {\ul 73.26$\pm$1.46} \\
 & 16 & 82.42$\pm$4.13 & 72.26$\pm$4.43 & 77.12$\pm$4.45 & 75.77$\pm$2.71 & 71.35$\pm$5.33 & 75.56$\pm$2.88 & 76.94$\pm$3.04 & 74.02$\pm$1.01 & \textbf{88.65$\pm$2.63} & 79.43$\pm$1.24 \\
 & 32 & 87.93$\pm$3.23 & 86.39$\pm$1.42 & 86.79$\pm$4.31 & 81.19$\pm$2.42 & 81.00$\pm$2.81 & 82.29$\pm$2.34 & 81.64$\pm$2.52 & 76.44$\pm$4.02 & \textbf{89.02$\pm$1.50} & 85.01$\pm$1.36 \\
 & 64 & 91.78$\pm$2.47 & 91.67$\pm$1.26 & \textbf{93.77$\pm$1.23} & 84.22$\pm$1.93 & 86.73$\pm$3.10 & 84.45$\pm$1.69 & 77.65$\pm$3.74 & 76.13$\pm$1.17 & 92.18$\pm$0.47 & 86.78$\pm$0.90 \\ \hline
Commun & 4 & 67.45$\pm$13.26 & 53.94$\pm$4.19 & 66.09$\pm$10.52 & 66.18$\pm$9.13 & N/A & 66.87$\pm$14.10 & N/A & N/A & N/A & \textbf{75.39$\pm$5.05} \\
ities & 8 & 73.73$\pm$5.45 & 66.65$\pm$4.50 & 71.16$\pm$4.61 & 72.69$\pm$3.79 & N/A & 76.36$\pm$4.55 & N/A & N/A & N/A & \textbf{76.59$\pm$1.25} \\
 & 16 & 72.55$\pm$4.83 & 68.01$\pm$1.97 & 71.66$\pm$4.81 & 73.09$\pm$2.84 & N/A & \textbf{77.29$\pm$2.56} & N/A & N/A & N/A & {\ul 76.25$\pm$0.64} \\
 & 32 & 73.65$\pm$3.04 & 72.72$\pm$3.64 & 73.25$\pm$6.11 & 76.09$\pm$3.40 & N/A & \textbf{78.60$\pm$2.59} & N/A & N/A & N/A & {\ul 76.85$\pm$1.38} \\
 & 64 & 77.78$\pm$5.16 & 78.33$\pm$4.52 & \textbf{79.05$\pm$3.80} & 77.79$\pm$3.43 & N/A & 79.00$\pm$1.61 & N/A & N/A & N/A & 78.62$\pm$1.39 \\ \hline
Credit-g & 4 & 52.68$\pm$4.46 & 50.00$\pm$0.00 & \textbf{57.00$\pm$10.75} & 48.92$\pm$4.60 & 54.00$\pm$7.34 & 48.80$\pm$6.76 & 52.99$\pm$4.08 & 54.33$\pm$6.54 & 51.90$\pm$9.40 & {\ul 55.94$\pm$1.10} \\
 & 8 & 55.52$\pm$8.88 & 52.22$\pm$4.90 & \textbf{59.84$\pm$7.33} & 55.26$\pm$3.92 & 52.58$\pm$11.27 & 54.50$\pm$8.25 & 52.43$\pm$4.36 & 52.90$\pm$5.79 & 56.42$\pm$12.89 & {\ul 57.42$\pm$3.10} \\
 & 16 & 58.26$\pm$5.17 & 56.23$\pm$4.37 & 58.42$\pm$8.36 & 59.22$\pm$11.38 & 58.91$\pm$8.04 & 57.63$\pm$7.58 & 55.29$\pm$4.80 & 51.65$\pm$4.02 & \textbf{60.38$\pm$14.03} & 56.60$\pm$2.22 \\
 & 32 & 67.85$\pm$5.78 & 65.33$\pm$6.28 & 57.48$\pm$5.73 & \textbf{72.60$\pm$7.18} & 66.27$\pm$5.06 & 63.24$\pm$5.47 & N/A & N/A & 68.64$\pm$3.86 & 61.79$\pm$10.25 \\
 & 64 & 72.77$\pm$9.29 & 70.79$\pm$2.34 & 68.53$\pm$4.99 & \textbf{73.12$\pm$6.23} & 68.95$\pm$6.14 & 70.97$\pm$4.95 & N/A & N/A & 70.80$\pm$4.09 & 66.43$\pm$2.90 \\ \hline
Diabetes & 4 & 57.09$\pm$18.84 & 50.00$\pm$0.00 & 52.50$\pm$7.77 & 62.35$\pm$7.48 & 56.28$\pm$13.01 & 64.22$\pm$6.78 & 71.71$\pm$5.31 & 63.96$\pm$3.32 & 70.42$\pm$3.69 & \textbf{80.28$\pm$0.75} \\
 & 8 & 65.52$\pm$13.18 & 50.86$\pm$22.03 & 65.34$\pm$8.84 & 64.69$\pm$13.33 & 69.08$\pm$9.68 & 67.39$\pm$12.92 & 72.21$\pm$2.07 & 65.47$\pm$3.95 & 64.30$\pm$5.88 & \textbf{79.38$\pm$1.66} \\
 & 16 & 73.44$\pm$0.52 & 65.69$\pm$6.54 & 65.69$\pm$6.33 & 71.86$\pm$3.16 & 73.69$\pm$3.21 & 73.79$\pm$6.48 & 71.64$\pm$5.05 & 66.71$\pm$0.76 & 67.34$\pm$2.79 & \textbf{80.15$\pm$1.35} \\
 & 32 & 73.95$\pm$3.32 & 72.97$\pm$3.77 & 71.27$\pm$8.04 & 72.91$\pm$3.09 & 75.22$\pm$3.21 & 76.70$\pm$4.55 & 73.32$\pm$1.59 & 66.97$\pm$1.75 & 69.74$\pm$4.41 & \textbf{80.06$\pm$1.18} \\
 & 64 & 74.52$\pm$1.59 & 72.56$\pm$3.17 & 76.92$\pm$2.39 & 74.44$\pm$4.13 & 77.82$\pm$3.49 & 78.64$\pm$3.32 & 70.22$\pm$4.09 & 69.27$\pm$6.15 & 71.56$\pm$4.55 & \textbf{80.91$\pm$1.62} \\ \hline
Heart & 4 & 70.54$\pm$3.83 & 50.00$\pm$0.00 & 70.85$\pm$2.02 & 59.38$\pm$3.42 & 67.33$\pm$15.29 & \textbf{88.27$\pm$3.32} & 60.76$\pm$4.00 & 68.19$\pm$11.17 & 59.74$\pm$4.49 & {\ul 75.66$\pm$4.59} \\
 & 8 & 78.12$\pm$10.59 & 55.88$\pm$3.98 & 79.43$\pm$4.28 & 74.35$\pm$6.93 & 77.89$\pm$2.34 & \textbf{88.78$\pm$2.38} & 65.46$\pm$3.77 & 69.85$\pm$10.82 & 70.14$\pm$7.91 & {\ul 79.46$\pm$2.16} \\
 & 16 & 83.02$\pm$3.70 & 78.62$\pm$7.14 & 83.45$\pm$3.95 & 83.66$\pm$5.91 & 81.45$\pm$5.05 & \textbf{89.13$\pm$2.10} & 67.00$\pm$7.83 & 68.39$\pm$11.73 & 81.72$\pm$3.92 & {\ul 83.71$\pm$1.88} \\
 & 32 & 84.84$\pm$3.53 & 87.11$\pm$1.22 & 88.77$\pm$2.36 & 88.45$\pm$2.43 & 88.00$\pm$2.34 & \textbf{89.65$\pm$3.04} & 71.94$\pm$3.88 & 71.90$\pm$9.07 & 87.43$\pm$2.32 & 87.19$\pm$3.66 \\
 & 64 & 89.74$\pm$3.30 & 89.99$\pm$3.82 & 89.74$\pm$2.62 & \textbf{90.97$\pm$1.55} & 90.20$\pm$2.57 & 89.62$\pm$3.16 & N/A & N/A & 89.78$\pm$2.59 & 88.08$\pm$4.11 \\ \hline
Myocardial & 4 & 51.25$\pm$3.85 & 50.00$\pm$0.00 & 51.91$\pm$4.49 & 47.70$\pm$4.10 & N/A & 52.77$\pm$2.01 & N/A & N/A & N/A & \textbf{52.87$\pm$3.44} \\
 & 8 & 55.34$\pm$1.11 & 55.63$\pm$2.92 & 52.77$\pm$5.83 & 49.37$\pm$3.41 & N/A & 55.40$\pm$4.41 & N/A & N/A & N/A & \textbf{56.22$\pm$1.64} \\
 & 16 & 60.00$\pm$5.16 & 56.55$\pm$12.22 & 54.16$\pm$4.53 & 54.31$\pm$1.42 & N/A & \textbf{61.22$\pm$3.45} & N/A & N/A & N/A & 55.32$\pm$9.15 \\
 & 32 & 58.63$\pm$0.96 & 57.31$\pm$5.31 & 46.43$\pm$5.02 & 53.52$\pm$0.74 & N/A & \textbf{60.76$\pm$1.58} & N/A & N/A & N/A & {\ul 60.02$\pm$4.02} \\
 & 64 & 57.04$\pm$1.94 & 56.18$\pm$2.85 & 55.00$\pm$4.73 & 54.41$\pm$2.00 & N/A & 59.79$\pm$0.56 & N/A & N/A & N/A & \textbf{61.47$\pm$3.91} \\ \hline
Cultivars & 4 & 53.45$\pm$10.79 & 50.00$\pm$0.00 & 48.29$\pm$10.29 & 46.99$\pm$6.33 & 49.80$\pm$15.90 & \textbf{57.10$\pm$8.66} & 51.38$\pm$2.48 & 54.28$\pm$3.73 & 54.39$\pm$5.61 & {\ul 55.63$\pm$5.24} \\
 & 8 & 56.22$\pm$11.87 & 52.60$\pm$6.31 & 50.41$\pm$4.93 & 51.76$\pm$9.99 & 54.72$\pm$9.35 & \textbf{57.26$\pm$9.52} & 51.68$\pm$4.43 & 51.48$\pm$3.85 & 52.86$\pm$6.13 & {\ul 56.97$\pm$5.08} \\
 & 16 & \textbf{60.35$\pm$4.23} & 56.87$\pm$2.50 & 55.89$\pm$3.81 & 57.06$\pm$9.27 & 54.92$\pm$8.32 & 60.09$\pm$7.64 & 54.31$\pm$6.12 & 57.44$\pm$3.53 & 56.97$\pm$2.22 & 57.19$\pm$5.30 \\
 & 32 & \textbf{60.94$\pm$12.64} & 50.81$\pm$6.53 & 52.64$\pm$2.13 & 57.29$\pm$6.98 & 55.08$\pm$10.85 & 60.48$\pm$6.51 & N/A & N/A & 58.50$\pm$2.65 & 59.62$\pm$7.43 \\
 & 64 & \textbf{65.62$\pm$5.53} & 49.82$\pm$3.43 & 53.87$\pm$5.37 & 59.02$\pm$4.58 & 56.25$\pm$7.36 & 61.07$\pm$6.77 & N/A & N/A & 60.32$\pm$2.60 & 59.14$\pm$4.79 \\ \hline
NHANES & 4 & 91.96$\pm$7.02 & 50.00$\pm$0.00 & 70.11$\pm$12.83 & 51.58$\pm$4.66 & 80.74$\pm$3.89 & 69.32$\pm$19.59 & 91.84$\pm$3.79 & 93.54$\pm$4.20 & \textbf{99.49$\pm$0.23} & 92.20$\pm$1.71 \\
 & 8 & 92.38$\pm$8.21 & 92.92$\pm$4.56 & 87.81$\pm$3.42 & 55.07$\pm$4.79 & 85.10$\pm$5.88 & 68.56$\pm$18.35 & 86.67$\pm$5.49 & 94.25$\pm$3.35 & \textbf{100.00$\pm$0.00} & 93.29$\pm$7.01 \\
 & 16 & 94.12$\pm$3.88 & 94.52$\pm$6.28 & 94.70$\pm$0.25 & 89.78$\pm$6.77 & 95.54$\pm$0.82 & 68.62$\pm$19.81 & 93.33$\pm$4.47 & 95.02$\pm$1.57 & \textbf{100.00$\pm$0.00} & {\ul 95.64$\pm$4.67} \\
 & 32 & 97.66$\pm$0.53 & 98.40$\pm$2.77 & 97.61$\pm$2.56 & 94.43$\pm$2.17 & 98.05$\pm$0.66 & 75.06$\pm$3.56 & 88.54$\pm$5.40 & 95.82$\pm$3.71 & \textbf{100.00$\pm$0.00} & 97.29$\pm$1.28 \\
 & 64 & 98.98$\pm$0.64 & 100.00$\pm$0.00 & 99.85$\pm$0.24 & 95.74$\pm$1.60 & 99.21$\pm$0.20 & 80.29$\pm$4.56 & N/A & N/A & \textbf{100.00$\pm$0.00} & 98.32$\pm$0.65 \\ \hline
Sequence & 4 & 66.80$\pm$4.98 & 50.00$\pm$0.00 & 67.81$\pm$1.70 & 67.08$\pm$2.75 & 71.93$\pm$4.17 & 60.12$\pm$6.51 & 93.01$\pm$1.84 & 91.00$\pm$4.25 & 75.38$\pm$5.21 & \textbf{98.60$\pm$0.66} \\
-type & 8 & 72.55$\pm$4.37 & 61.01$\pm$10.96 & 64.27$\pm$7.56 & 60.53$\pm$6.91 & 79.55$\pm$6.71 & 60.58$\pm$9.58 & 94.29$\pm$1.64 & 90.18$\pm$3.73 & 79.91$\pm$2.13 & \textbf{98.70$\pm$1.44} \\
 & 16 & 83.10$\pm$6.86 & 80.31$\pm$5.79 & 72.58$\pm$3.51 & 73.14$\pm$5.29 & 89.59$\pm$2.33 & 65.12$\pm$5.28 & 93.46$\pm$0.72 & 92.09$\pm$2.90 & 86.97$\pm$1.97 & \textbf{99.29$\pm$0.78} \\
 & 32 & 93.09$\pm$2.32 & 86.09$\pm$2.54 & 86.19$\pm$4.70 & 70.12$\pm$5.27 & 96.12$\pm$2.31 & 65.67$\pm$3.95 & 96.05$\pm$2.43 & 96.08$\pm$3.53 & 96.98$\pm$1.22 & \textbf{99.67$\pm$0.12} \\
 & 64 & 94.63$\pm$1.43 & 90.07$\pm$2.64 & 90.60$\pm$2.76 & 67.54$\pm$3.40 & 97.74$\pm$0.34 & 68.56$\pm$0.72 & 98.20$\pm$2.11 & 98.06$\pm$2.62 & 99.01$\pm$0.26 & \textbf{99.54$\pm$0.48} \\ \hline
Solution & 4 & 72.71$\pm$1.96 & 50.00$\pm$0.00 & 67.70$\pm$7.66 & 68.48$\pm$14.97 & 71.19$\pm$3.90 & 64.52$\pm$8.41 & 73.53$\pm$1.94 & 80.26$\pm$5.11 & 73.75$\pm$3.90 & \textbf{100.00$\pm$0.00} \\
-mix & 8 & 82.91$\pm$7.74 & 58.31$\pm$19.17 & 71.58$\pm$10.17 & 56.67$\pm$7.64 & 82.42$\pm$3.72 & 72.04$\pm$10.34 & 76.34$\pm$4.26 & 83.19$\pm$3.70 & 76.57$\pm$4.80 & \textbf{99.81$\pm$0.08} \\
 & 16 & 83.87$\pm$1.90 & 68.09$\pm$1.76 & 76.47$\pm$0.56 & 70.37$\pm$13.96 & 84.39$\pm$0.83 & 71.82$\pm$8.79 & 82.15$\pm$5.01 & 88.10$\pm$4.72 & 84.28$\pm$5.15 & \textbf{99.67$\pm$0.13} \\
 & 32 & 91.29$\pm$1.43 & 85.52$\pm$8.05 & 86.74$\pm$6.63 & 76.33$\pm$10.57 & 92.44$\pm$3.47 & 73.08$\pm$11.62 & 82.44$\pm$5.43 & 88.15$\pm$2.99 & 85.39$\pm$2.09 & \textbf{100.00$\pm$0.00} \\
 & 64 & 93.18$\pm$1.48 & 89.36$\pm$1.88 & 92.44$\pm$1.26 & 69.78$\pm$2.76 & 95.44$\pm$2.63 & 80.05$\pm$2.13 & N/A & N/A & 92.40$\pm$3.06 & \textbf{98.89$\pm$1.57} \\ \bottomrule
\end{tabular}}
\end{table}

\newpage
\subsection{Ablation Study}
\begin{table}[h!]
\caption{Ablation study results, showing AUC across 13 datasets.}
\centering
\scalebox{0.74}{
\begin{tabular}{@{}lcccccc@{}}
\toprule
Data & Shot & \model{} & -Tuning & -Ensemble & -Description & -Reasoning \\ \midrule
Adult & 4 & 86.68$\pm$0.86 & 87.19$\pm$0.45 & 84.49$\pm$1.79 & 86.54$\pm$1.16 & 85.33$\pm$0.75 \\
 & 8 & 87.89$\pm$0.06 & 87.34$\pm$0.89 & 84.54$\pm$4.93 & 87.54$\pm$0.39 & 86.13$\pm$1.21 \\
 & 16 & 87.54$\pm$0.50 & 87.54$\pm$0.40 & 84.75$\pm$2.88 & 87.66$\pm$0.56 & 86.55$\pm$0.39 \\
 & 32 & 87.09$\pm$0.58 & 87.56$\pm$0.27 & 84.65$\pm$2.20 & 87.71$\pm$0.75 & 85.96$\pm$1.10 \\
 & 64 & 87.77$\pm$0.31 & 87.77$\pm$0.39 & 84.94$\pm$2.00 & 87.76$\pm$0.29 & 86.73$\pm$0.15 \\ \hline
Bank & 4 & 70.45$\pm$3.69 & 66.19$\pm$1.88 & 63.88$\pm$7.37 & 69.46$\pm$1.54 & 68.77$\pm$3.64 \\
 & 8 & 75.85$\pm$6.66 & 68.78$\pm$1.07 & 67.94$\pm$7.03 & 75.30$\pm$3.23 & 70.33$\pm$1.55 \\
 & 16 & 78.41$\pm$1.08 & 72.73$\pm$0.41 & 68.74$\pm$11.83 & 77.30$\pm$0.41 & 79.93$\pm$1.54 \\
 & 32 & 78.37$\pm$4.50 & 69.89$\pm$5.35 & 69.08$\pm$10.89 & 78.74$\pm$2.97 & 77.97$\pm$6.66 \\
 & 64 & 81.18$\pm$6.17 & 72.27$\pm$2.09 & 72.03$\pm$8.81 & 79.88$\pm$3.05 & 79.03$\pm$6.70 \\ \hline
Blood & 4 & 68.34$\pm$7.48 & 69.31$\pm$5.92 & 61.17$\pm$9.30 & 57.69$\pm$6.45 & 65.42$\pm$2.51 \\
 & 8 & 70.37$\pm$3.23 & 70.41$\pm$3.21 & 64.76$\pm$7.29 & 68.66$\pm$3.99 & 68.77$\pm$4.12 \\
 & 16 & 70.07$\pm$5.19 & 69.99$\pm$4.66 & 65.56$\pm$8.29 & 72.01$\pm$1.92 & 68.66$\pm$3.18 \\
 & 32 & 71.13$\pm$4.38 & 70.57$\pm$2.81 & 67.52$\pm$5.37 & 71.25$\pm$2.98 & 69.50$\pm$2.57 \\
 & 64 & 71.04$\pm$4.36 & 71.55$\pm$4.51 & 67.24$\pm$4.78 & 70.84$\pm$1.35 & 69.67$\pm$2.17 \\ \hline
Car & 4 & 72.69$\pm$1.52 & 71.36$\pm$1.31 & 69.89$\pm$3.75 & 69.17$\pm$2.84 & 69.27$\pm$0.05 \\
 & 8 & 73.26$\pm$1.46 & 72.52$\pm$1.55 & 70.54$\pm$4.32 & 71.76$\pm$5.59 & 72.76$\pm$4.32 \\
 & 16 & 79.43$\pm$1.24 & 74.64$\pm$0.40 & 74.65$\pm$5.46 & 78.48$\pm$1.18 & 81.82$\pm$2.04 \\
 & 32 & 85.01$\pm$1.36 & 74.89$\pm$0.97 & 81.20$\pm$5.50 & 85.72$\pm$0.18 & 85.58$\pm$2.11 \\
 & 64 & 86.78$\pm$0.90 & 74.16$\pm$0.27 & 81.81$\pm$6.96 & 86.55$\pm$1.49 & 87.10$\pm$1.49 \\ \hline
Communities & 4 & 75.39$\pm$5.05 & 77.42$\pm$1.16 & 66.26$\pm$7.07 & 77.22$\pm$3.08 & 71.97$\pm$9.87 \\
 & 8 & 76.59$\pm$1.25 & 75.92$\pm$1.69 & 65.89$\pm$6.33 & 76.38$\pm$0.93 & 69.63$\pm$6.74 \\
 & 16 & 76.25$\pm$0.64 & 74.05$\pm$1.97 & 66.33$\pm$6.95 & 76.42$\pm$2.52 & 75.13$\pm$3.61 \\
 & 32 & 76.85$\pm$1.38 & 71.64$\pm$4.71 & 66.53$\pm$5.82 & 77.61$\pm$1.11 & 77.39$\pm$1.72 \\
 & 64 & 78.62$\pm$1.39 & 75.80$\pm$0.29 & 70.50$\pm$4.94 & 78.87$\pm$1.81 & 79.10$\pm$2.38 \\ \hline
Credit-g & 4 & 55.94$\pm$1.10 & 55.70$\pm$1.93 & 53.56$\pm$5.24 & 53.64$\pm$2.03 & 50.76$\pm$0.87 \\
 & 8 & 57.42$\pm$3.10 & 53.51$\pm$4.26 & 53.08$\pm$7.09 & 54.13$\pm$1.76 & 49.02$\pm$2.68 \\
 & 16 & 56.60$\pm$2.22 & 55.48$\pm$3.48 & 54.24$\pm$7.03 & 54.96$\pm$8.64 & 56.99$\pm$3.67 \\
 & 32 & 61.79$\pm$10.25 & 53.30$\pm$3.76 & 55.51$\pm$8.61 & 62.14$\pm$4.61 & 62.22$\pm$2.49 \\
 & 64 & 66.43$\pm$2.90 & 57.57$\pm$2.88 & 57.76$\pm$6.91 & 69.09$\pm$4.21 & 63.76$\pm$6.43 \\ \hline
Diabetes & 4 & 80.28$\pm$0.75 & 78.78$\pm$1.21 & 73.95$\pm$9.66 & 78.88$\pm$1.90 & 77.33$\pm$2.44 \\
 & 8 & 79.38$\pm$1.66 & 79.81$\pm$1.80 & 75.95$\pm$5.59 & 78.41$\pm$2.63 & 77.09$\pm$1.61 \\
 & 16 & 80.15$\pm$1.35 & 78.76$\pm$0.47 & 77.44$\pm$2.95 & 79.58$\pm$0.12 & 78.63$\pm$1.60 \\
 & 32 & 80.06$\pm$1.18 & 78.67$\pm$1.93 & 78.35$\pm$2.27 & 79.56$\pm$1.37 & 78.54$\pm$0.86 \\
 & 64 & 80.91$\pm$1.62 & 79.48$\pm$1.61 & 79.07$\pm$2.31 & 80.39$\pm$1.17 & 78.90$\pm$1.20 \\ \hline
Heart & 4 & 75.66$\pm$4.59 & 74.67$\pm$4.22 & 65.91$\pm$16.49 & 68.01$\pm$8.43 & 48.03$\pm$8.08 \\
 & 8 & 79.46$\pm$2.16 & 69.66$\pm$0.74 & 64.57$\pm$19.44 & 79.06$\pm$5.32 & 78.58$\pm$9.07 \\
 & 16 & 83.71$\pm$1.88 & 76.57$\pm$2.11 & 71.58$\pm$16.06 & 83.35$\pm$4.09 & 80.39$\pm$6.28 \\
 & 32 & 87.19$\pm$3.66 & 77.37$\pm$4.39 & 76.61$\pm$14.45 & 86.20$\pm$3.27 & 85.37$\pm$3.33 \\
 & 64 & 88.08$\pm$4.11 & 81.32$\pm$6.06 & 82.72$\pm$9.23 & 87.69$\pm$2.33 & 87.55$\pm$2.04 \\ \hline
Myocardial & 4 & 52.87$\pm$3.44 & 54.14$\pm$2.89 & 51.10$\pm$6.43 & 56.12$\pm$7.20 & 49.47$\pm$4.93 \\
 & 8 & 56.22$\pm$1.64 & 55.42$\pm$2.57 & 52.31$\pm$6.14 & 52.84$\pm$4.41 & 51.42$\pm$8.06 \\
 & 16 & 55.32$\pm$9.15 & 54.73$\pm$7.07 & 51.73$\pm$7.61 & 56.00$\pm$2.51 & 50.03$\pm$13.37 \\
 & 32 & 60.02$\pm$4.02 & 55.20$\pm$6.38 & 52.29$\pm$7.07 & 62.20$\pm$4.95 & 46.40$\pm$11.98 \\
 & 64 & 61.47$\pm$3.91 & 57.19$\pm$1.25 & 53.04$\pm$6.26 & 57.10$\pm$6.05 & 59.60$\pm$1.42 \\ \hline
Cultivars & 4 & 55.63$\pm$5.24 & 56.58$\pm$3.34 & 54.41$\pm$4.33 & 52.98$\pm$2.44 & 53.73$\pm$5.23 \\
 & 8 & 56.97$\pm$5.08 & 56.69$\pm$4.64 & 55.29$\pm$5.51 & 54.46$\pm$3.86 & 52.86$\pm$3.86 \\
 & 16 & 57.19$\pm$5.30 & 55.40$\pm$4.71 & 55.06$\pm$6.98 & 55.37$\pm$6.77 & 56.71$\pm$7.72 \\
 & 32 & 59.62$\pm$7.43 & 52.96$\pm$5.95 & 53.33$\pm$6.10 & 53.74$\pm$7.43 & 56.15$\pm$7.55 \\
 & 64 & 59.14$\pm$4.79 & 56.84$\pm$4.11 & 55.74$\pm$6.08 & 54.46$\pm$4.79 & 54.65$\pm$4.07 \\ \hline
NHANES & 4 & 92.20$\pm$1.71 & 80.73$\pm$3.89 & 84.89$\pm$7.12 & 93.73$\pm$2.31 & 89.85$\pm$2.99 \\
 & 8 & 93.29$\pm$7.01 & 86.24$\pm$6.20 & 86.94$\pm$7.90 & 92.62$\pm$3.53 & 91.26$\pm$6.38 \\
 & 16 & 95.64$\pm$4.67 & 88.66$\pm$6.17 & 88.70$\pm$9.11 & 96.63$\pm$2.76 & 95.29$\pm$3.48 \\
 & 32 & 97.29$\pm$1.28 & 83.97$\pm$3.74 & 92.88$\pm$5.72 & 99.04$\pm$0.79 & 96.89$\pm$2.47 \\
 & 64 & 98.32$\pm$0.65 & 83.64$\pm$3.77 & 95.21$\pm$4.25 & 97.41$\pm$0.58 & 96.15$\pm$1.54 \\ \hline
Sequence-type & 4 & 98.60$\pm$0.66 & 94.35$\pm$3.98 & 91.06$\pm$7.78 & 99.00$\pm$0.29 & 97.89$\pm$0.54 \\
 & 8 & 98.70$\pm$1.44 & 94.32$\pm$5.25 & 91.81$\pm$7.20 & 98.96$\pm$0.40 & 99.14$\pm$0.10 \\
 & 16 & 99.29$\pm$0.78 & 97.75$\pm$2.47 & 89.45$\pm$11.66 & 99.68$\pm$0.28 & 99.62$\pm$0.54 \\
 & 32 & 99.67$\pm$0.12 & 93.92$\pm$5.13 & 90.11$\pm$8.90 & 99.33$\pm$0.83 & 99.36$\pm$0.57 \\
 & 64 & 99.54$\pm$0.48 & 98.33$\pm$0.24 & 94.18$\pm$6.92 & 99.56$\pm$0.39 & 99.79$\pm$0.36 \\ \hline
Solution-mix & 4 & 100.00$\pm$0.00 & 100.00$\pm$0.00 & 94.08$\pm$11.97 & 99.41$\pm$0.51 & 91.58$\pm$6.68 \\
 & 8 & 99.81$\pm$0.08 & 99.17$\pm$0.13 & 81.12$\pm$14.71 & 99.52$\pm$0.45 & 92.07$\pm$1.52 \\
 & 16 & 99.67$\pm$0.13 & 99.50$\pm$0.07 & 84.64$\pm$22.27 & 98.44$\pm$1.74 & 90.06$\pm$6.66 \\
 & 32 & 100.00$\pm$0.00 & 99.42$\pm$0.12 & 80.07$\pm$21.38 & 97.11$\pm$2.51 & 91.32$\pm$7.06 \\
 & 64 & 98.89$\pm$1.57 & 98.83$\pm$1.65 & 84.73$\pm$19.43 & 99.48$\pm$0.34 & 93.85$\pm$1.18 \\ \bottomrule
\end{tabular}}
\end{table}

\end{document}